\documentclass[journal]{IEEEtran}
\usepackage{graphicx}
\usepackage{algorithmic}
\usepackage[ruled]{algorithm2e}
\usepackage{xspace}
\usepackage{eufrak}
\usepackage{eucal}
\usepackage{makecell}
\usepackage{array}
\usepackage{multirow}
\usepackage{amsmath}
\usepackage{amssymb}
\usepackage{tabularx}
\usepackage{booktabs}

\usepackage{pifont}

\usepackage[dvipsnames]{xcolor}

\newcommand\blue[1]{\color{blue}{#1}}
\newcommand{\red}[1]{{\color{red}#1}}

\usepackage{bm}      % \bm
\newcommand{\revise}[1]{\textcolor{black}{#1}}

\def\0{{\bf 0}}
\def\1{{\bf 1}}

\usepackage{cite}
\ifCLASSINFOpdf
\else
\fi
\usepackage{amsmath}
\usepackage{algorithmic}
\usepackage{array}
\begin{document}
\title{PromptDLA: A Domain-aware Prompt Document Layout Analysis Framework with Descriptive Knowledge as a Cue}

\author{ Zirui Zhang\textsuperscript{*},Yaping Zhang\textsuperscript{*}, Lu Xiang, Yang Zhao, Feifei Zhai, Yu Zhou and Chengqing Zong,~\IEEEmembership{Fellow,~IEEE}% <-this % stops a space
\thanks{Y.~Zhang, L.~Xiang, Y.~Zhao, F.~Zhai, Y.~Zhou and C.~Zong are with the Institute of Automation, Chinese Academy of Sciences, and the University of the Chinese Academy of Sciences. Y.~Zhou is also with the Fanyu AI Laboratory, Zhongke Fanyu Technology Co., Ltd. Z.~Zhang is with Columbia University; this work was done while at the Fanyu AI Laboratory.
Email: zz3093@columbia.edu, \{yaping.zhang, lu.xiang, yang.zhao, yzhou, cqzong\}@nlpr.ia.ac.cn. Y.~Zhang is the corresponding author.}
\thanks{Y.~Zhang and Z.~Zhang contributed equally and are co-first authors.}
\thanks{This work is supported by the National Natural Science Foundation of China under Grant 62476275 and 62106265, and by the Young Scientists Fund of the State Key Laboratory of Multimodal Artificial Intelligence Systems.}}


\markboth{Journal of \LaTeX\ Class Files,~Vol.~14, No.~8, August~2015}%
{Shell \MakeLowercase{\textit{et al.}}: Bare Demo of IEEEtran.cls for IEEE Journals}



\maketitle

\begin{abstract}
Document Layout Analysis (DLA) is crucial for document artificial intelligence and has recently received increasing attention, resulting in an influx of large-scale public DLA datasets. 
Existing work often combines data from various domains in recent public DLA datasets to improve the generalization of DLA. 
However, directly merging these datasets for training often results in suboptimal model performance, as it overlooks the different layout structures inherent to various domains. These variations include different labeling styles, document types, and languages.
This paper introduces PromptDLA, a domain-aware Prompter for Document Layout Analysis that effectively leverages descriptive knowledge as cues to integrate domain priors into DLA. The innovative PromptDLA features a unique domain-aware prompter that customizes prompts based on the specific attributes of the data domain. These prompts then serve as cues that direct the DLA toward critical features and structures within the data, enhancing the model's ability to generalize across varied domains. Extensive experiments show that our proposal achieves state-of-the-art performance among DocLayNet, PubLayNet, M6Doc, and D$^4$LA. Our code is available on GitHub.\footnote{https://github.com/Zirui00/PromptDLA.git}
%However, the distribution differences between these domains are substantial; %sometimes, there are even conflictive label styles. 
%Directly merging these data for training leads to sub-optimal model performance. 
% This paper proposes a simple yet effective domain-aware Prompter for Document Layout Analysis (PromptDLA) to incorporate domain prior into DLA through domain prompts. The proposed novel PromptDLA features a unique domain-aware prompter that customizes prompts according to the specific attributes of the data domain. Then, these prompts can guide the DLA toward essential features and structures in the data, leading to better generalization of the DLA across different domains. 
\end{abstract}


% \fi
\IEEEpeerreviewmaketitle

\section{Introduction}
\label{sec:intro}
% What is DLA and why is it important?
%TODO: 添加2025年 DLA 文献，同时加入LLM 对DLA应用的文献参考
\quad Document Layout Analysis (DLA) aims to distinguish the physical or logical layout structure of documents~\cite{cai2018cascade,bao2021beit,zhang2022multimodal, bi2022srrv, banerjee2023swindocsegmenter,HRDocAAAI2023}, identifying areas characterized by elements such as text, image, and table. {It is fundamental to modern document artificial intelligence, significantly influencing subsequent document understanding tasks such as information extraction and digital transformation} \cite{ma2023multi, he2023icl, luo2024layoutllm, wang2024knowledge,liang2024document, hu2024mplug, zhang2023layoutdit}.

\begin{figure}[!t]
	\centering
	\includegraphics[width=1\columnwidth]{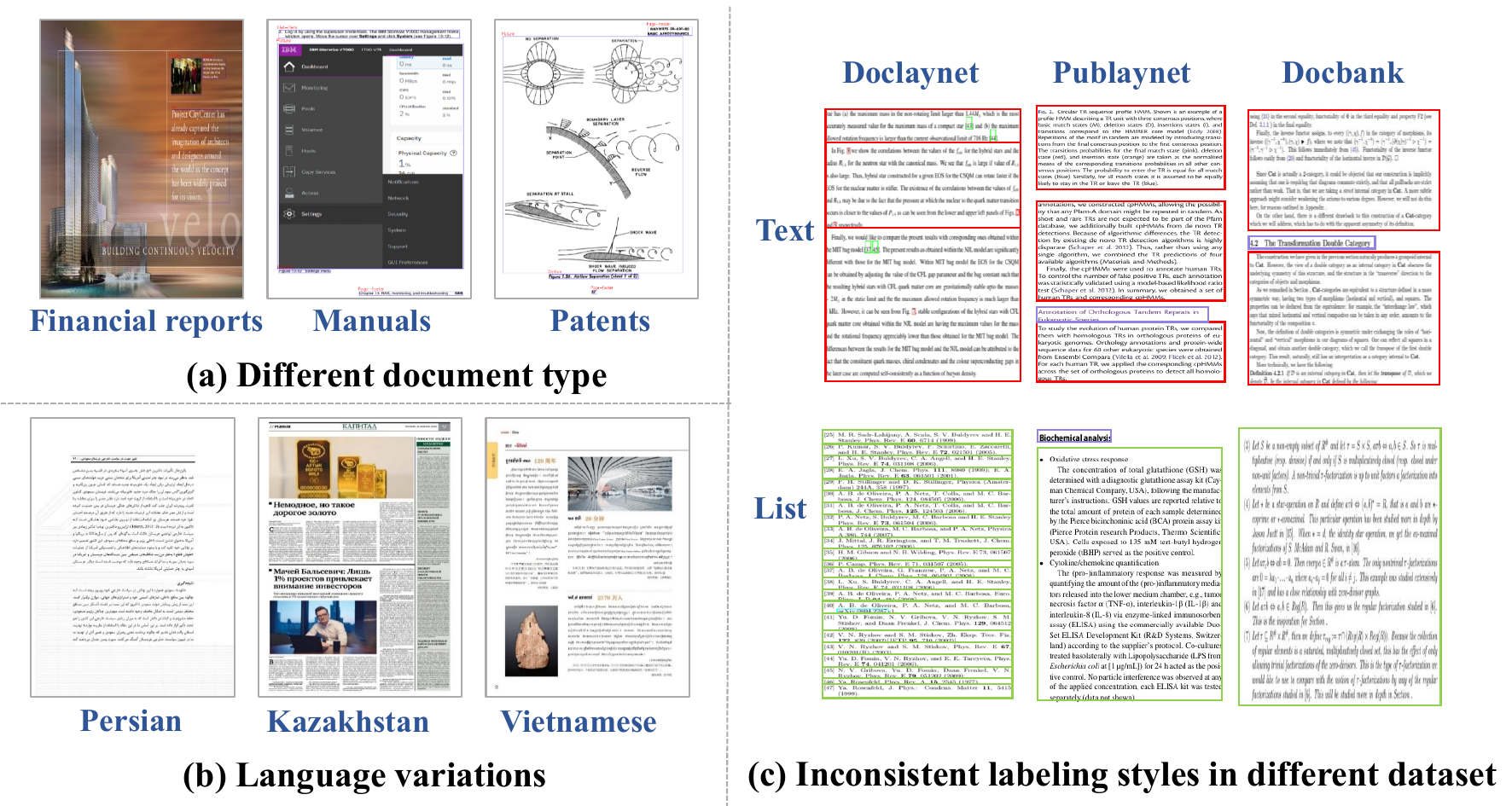}
	\caption{Examples of different domain differences across (\textbf{a}) Different document types \revise{ caused variations in layout structure and element distribution (financial report, manual, patent)} (\textbf{b}) different language types, and (\textbf{c})Inconsistent labeling styles. Note that the \textcolor{red}{"text"} and \textcolor{green}{"list"} items in DocLayNet are labeled as smaller individual units while they are integrated as a whole in DocBank.}
	\label{vis-exmaples-new}
\end{figure}

\begin{figure*}[t]
	\centering
	\includegraphics[width=0.78\linewidth]{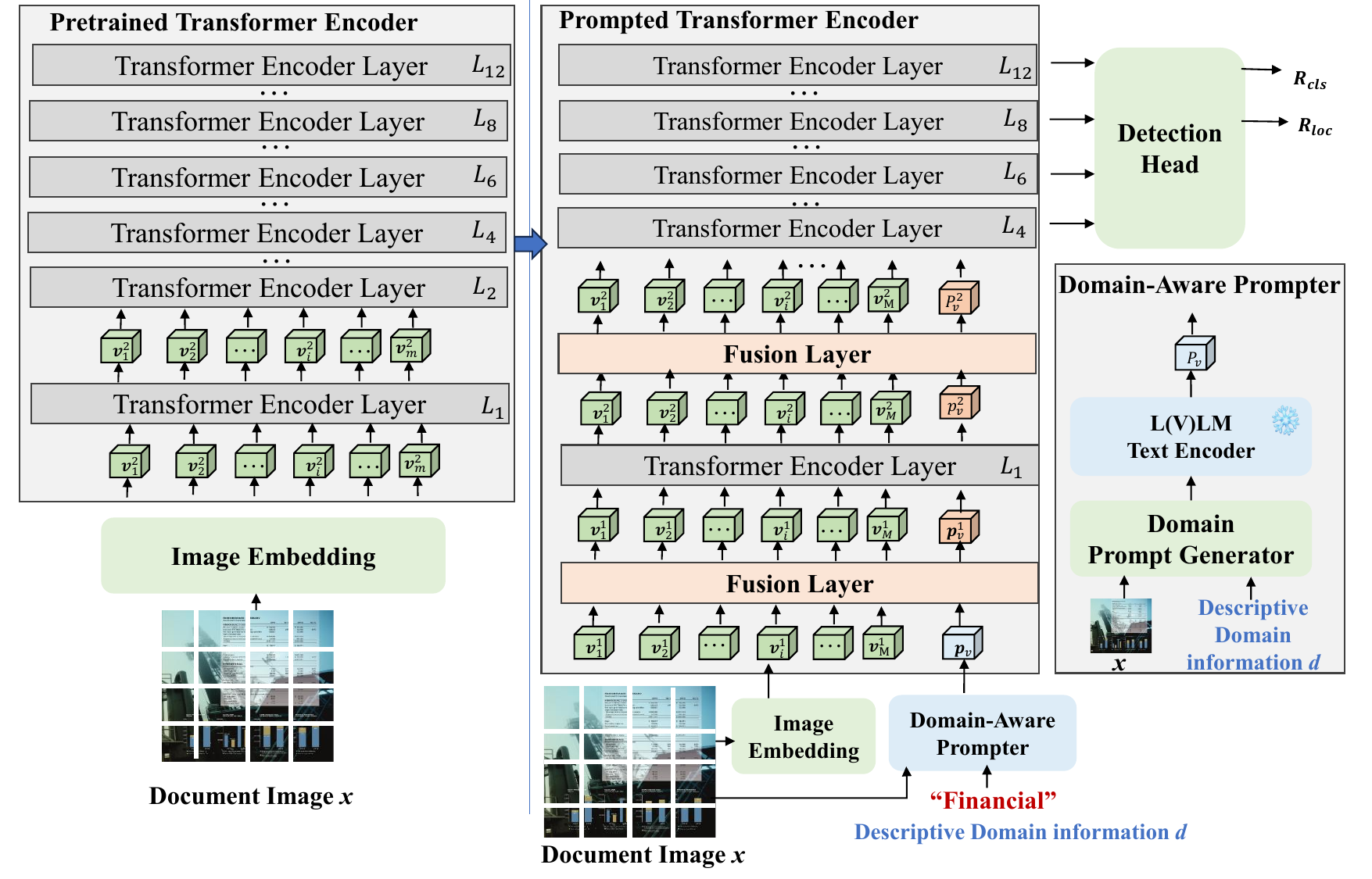} 
\caption{Overview of the PromptDLA method for domain-aware layout prediction. A Domain-Aware Prompter encodes domain information into a prompt vector, which is prepended to the sequence of image patch embeddings. This augmented input is processed by a vision backbone. Multi-scale features are extracted from the backbone and refined by a FPN before being passed to a detection head for final layout prediction.}
	\label{framework}
\end{figure*}

{With growing research interest in DLA, large-scale datasets have emerged, such as} PubLayNet~\cite{zhong2019PubLayNet}, DocBank~\cite{li2020docbank}, DocLayNet~\cite{pfitzmann2022DocLayNet}, M6Doc~\cite{cheng2023m6doc} and D$^4$LA~\cite{da2023vision}. 
To enhance generalizability in real-world scenarios, recent DLA datasets such as DocLayNet~\cite{pfitzmann2022DocLayNet}, M6Doc~\cite{cheng2023m6doc}, and D$^4$LA~\cite{da2023vision} have increased document diversity by combining data from various domains, including finance, law, and patents. {However, merging data from these diverse domains introduces substantial distribution differences, both across and within datasets.} Fig.~\ref{vis-exmaples-new} illustrates three critical yet overlooked domain differences encountered:
\begin{itemize}
	\item \textbf{Different document types}. Document images from different types exhibit unique layout structures and label distributions. Fig.~\ref{vis-exmaples-new}(a) visually contrasts the typical visual features of three document types: Financial reports (left) are characterized by well-designed color images and text-image overlays; manuals (center) frequently use functional screenshots of software interfaces to present data and controls; while patents (right) are dominated by structured, black-and-white technical line drawings with concise labels. These significant differences in layout highlight the need to incorporate domain priors into DLA.
    \item \textbf{Different languages}. Document images from various countries exhibit unique layout structures influenced by their respective languages. Fig.~\ref{vis-exmaples-new}(b) demonstrates how language affects the layout of the document. Persian documents predominantly feature dense blocks of text arranged in continuous paragraphs. In contrast, documents from Kazakhstan integrate numerous small paragraphs interspersed with images, creating a visually diverse page layout. These variations highlight the correlation between language and document layout, as shown in Fig.~\ref{vis-exmaples-new}.  
	\item \textbf{Inconsistent labeling styles}. Different datasets often adopt disparate annotation guidelines leading to conflict labeling style, even for semantically similar elements. As shown in Fig.~\ref{vis-exmaples-new}(c), DocLayNet annotates individual list items, whereas DocBank and PubLayNet group entire lists into single bounding boxes. Similarly, paragraph segmentation varies significantly across datasets. Such inconsistencies create conflicts during joint training and pose obstacles to building scalable, unified models.
    
\end{itemize}

\begin{figure}[!t]
	\centering
	\includegraphics[width=0.95\columnwidth]{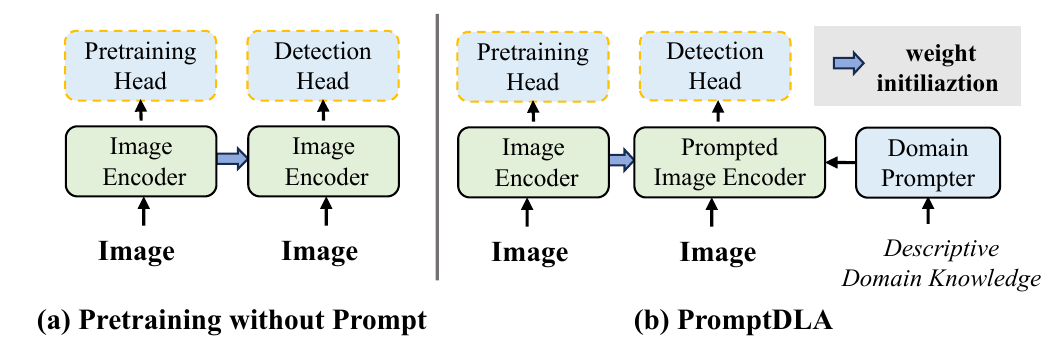}
	\caption{Comparison with Pre-training Paradigms in DLA}
	\label{famecomp}
\end{figure}

\begin{table*}[h]\centering
	\setlength{\tabcolsep}{2pt}
	\renewcommand{\arraystretch}{1}
	\caption{The Detail of document types in different DLA datasets.}
	\label{tab:datasets}
	% \resizebox{\columnwidth}{!}
	{
		\begin{tabular}{ccccccc}
			\toprule[1.5pt]
			\textbf{Dataset} & \textbf{Document Type} & \textbf{A.M.} & \textbf{Format} & \textbf{\#Class} & \textbf{Language}   & \textbf{\#Images} \\
			\midrule
			PubLayNet~\cite{zhong2019PubLayNet}	& Articles & Automatic & PDF    & 5  & English  &364232	 \\
			\hline
			%	\hline
			%	DocBank~\cite{li2020docbank} 	& Articles    & 13 & English	 &400,000/50,000 \\
			%	\hline
			DocLayNet~\cite{pfitzmann2022DocLayNet}    & \begin{tabular}[c]{@{}c@{}}Financial Reports, Manuals, \\ Scientific Articles, Laws \& Regulations, \\ Patents, Government Tenders.\end{tabular} & Manual & PDF  & 11 & \begin{tabular}[c]{@{}c@{}}English, German, \\ French, Japanese\end{tabular}	 &{80863} \\
			\hline
			M6Doc~\cite{cheng2023m6doc} & \begin{tabular}[c]{@{}c@{}} Scientific articles, Textbooks,  Books,\\ Test papers, Magazines, Newspapers, Notes\end{tabular} & Manual & PDF, Scanned, Photographed	   & 74  &English,Chinese	 &{9080} \\
			\hline
			D$^4$LA~\cite{da2023vision}    & \begin{tabular}[c]{@{}c@{}} Scientific report, Email, Form, Invoice, Letter,\\ Specification, News article, Presentation, Resume,\\ Scientific publication, Budget, Memo  \end{tabular} & Manual & PDF, Scanned,	   & 27 &English	  &11092 \\
			%					\hline
			%					{MLDLA}(\textbf{Ours})  & Government Reports     & 13 & \begin{tabular}[c]{@{}c@{}}Kazakh, Persian, \\Hindi, Khmer, Lao, \\ Trukish, Vietnamese \end{tabular}	   &\checkmark &170,000/3,200 \\
            \hline
            \blue{MLDLA(ours)}                          & \begin{tabular}[c]{@{}l@{}}Magazine, Newspaper, Government Reports\end{tabular} & Manual & PDF, Scanned, Photographed	  & 5 & \begin{tabular}[c]{@{}l@{}}Hindi, Kazakhstan, Vietnam,\\ Turkey, Persia, Laos, Khmer\end{tabular} &17505             \\
			\bottomrule[1pt]
	\end{tabular}}
	% \vspace{-4mm}
\end{table*}

These overlooked domain-related discrepancies can hinder the learning process and reduce the generalization capability of DLA models trained on combined datasets.A promising direction to mitigate these issues involves endowing DLA models with the capacity to adapt their analysis based on the specific characteristics of the input document's domain.
Recent studies in large vision-language models (LVLMs) like CLIP~\cite{radford2021learning} and large language models (LLMs) like LLaMA~\cite{touvron2023llama} have demonstrated the efficacy of prompt engineering for conditioning model behavior on domain-specific contexts across various tasks~\cite{Zhou_2022_CVPR, Lee_2023_ICCV, wei2022chain, liu2024large, chen2023unleashing, wang2024chain}. 
Inspired by this, we propose a novel framework for Domain-aware \textbf{Prompt} \textbf{D}ocument \textbf{L}ayout \textbf{A}nalysis, named \textbf{PromptDLA}. Unlike traditional pretraining-based DLA methods, PromptDLA integrates domain priors directly into the analysis process using LLMs like LLAMA~\cite{touvron2023llama} or LVLMs such as CLIP~\cite{radford2021learning}, BLIP2~\cite{li2023blip}. Central to our approach is a prompted transformer encoder, fine-tuned with a novel domain-aware prompter (depicted in Fig.~\ref{framework}). This prompter uses descriptive knowledge from the domain information of corresponding images as cues, guiding the transformer encoder to recognize and adapt to the variability across different domains effectively.
We evaluate our method on the DocLayNet, M6Doc, D$^4$LA, and the integration of PubLayNet and DocLayNet datasets. The experimental results have shown that the domain-aware model has outperformed the current state-of-the-art method. In addition, due to the predominance of English in existing datasets, we have introduced a multilingual DLA dataset—MLDLA, which contains document images in seven different languages. Detailed experiments have demonstrated that our method can effectively generalize across scenarios where language serves as domain information.
The contributions of this paper are summarized as follows:
\begin{itemize}
	\item A novel domain-aware DLA framework, named PromptDLA, is proposed that explicitly introduces domain knowledge to DLA, enabling models to better handle variability across a variety of document domains.
   \item  A unique and modular domain-aware prompter is proposed to generate customized prompts reflecting specific data attributes. It integrates easily with various backbone architectures, such as CNNs, ViTs, and Swin Transformers, using prompts derived from either human knowledge or LLMs generation.
	\item We conduct extensive experiments demonstrating the effectiveness of PromptDLA across multiple domain information types and datasets, including the newly introduced MLDLA benchmark, achieving state-of-the-art results.
\end{itemize} 

\section{Related Work}
\subsection{Document Layout Analysis}
The current literature on document analysis has redefined document understanding as a broad term that covers various problems and tasks related to document intelligence systems. 
Based on whether pre-training on large-scale unlabeled document images can divide the Document Layout Analysis method into two categories: traditional object detection frameworks and document pre-training methods.
Traditional object detection frameworks, such as Faster-RCNN\cite{ren2015faster}, Mask-RCNN\cite{he2017mask}, and YOLO\cite{redmon2016you}, typically train models directly on DLA datasets and may occasionally use ImageNet\cite{deng2009imagenet} pre-trained weights. 
In comparison, another approach involves training the transformer using the self-supervised method on a vast unlabeled document dataset and utilizing the pretrained transformer as the backbone of a two-stage Object Detection Framework. LayoutLMv3\cite{huang2022layoutlmv3} and Structextv2\cite{yu2023structextv2} are critical works in this area. It employs the multi-modal pre-training method, including Mask Image Modeling, Mask Language Modeling, and Word-Patch Alignment, using both the textual information on the image and the image itself as input to the transformer encoder. Other notable works in this area are DiT\cite{li2022dit}, DocFormer\cite{appalaraju2021docformer}, UniDoc\cite{gu2021unidoc}, and Self-Docseg\cite{maity2023selfdocseg}, which relies solely on a vision model, closely aligning with the approach of BEiT\cite{bao2021beit}, and directly applies a general pre-training framework to learn from large-scale document image data. Traditional approaches focus on enhancing the model's performance.
As depicted in Fig.~\ref{famecomp}, PromptDLA diverges from conventional pre-training methods by directly injecting domain knowledge into the DLA framework. Instead of implicitly learning domain features, our approach leverages descriptive prompts from LLMs, offering a more flexible and resource-efficient paradigm for domain adaptation.

\subsection{Prompt Engineering}
Prompt engineering has significantly improved the adaptability of LLMs and LVLMs for specialized tasks. By introducing learnable tokens into vision transformers (ViTs), models can effectively focus on task-relevant features with minimal structural changes~\cite{rezaei2024learning}. This strategy is similar to advances in multimodal learning, exemplified by CLIP, which jointly leverages text and image data to enhance robustness and versatility across various visual recognition tasks, particularly in zero-shot or few-shot settings~\cite{radford2021learning}. Furthermore, prompt engineering more generally utilizes carefully designed prompts to effectively elicit contextually appropriate outputs from pre-trained models, greatly reducing the need for extensive retraining~\cite{brown2020language}.

\begin{figure}[!t]
	\centering
	\includegraphics[width=0.8\columnwidth]{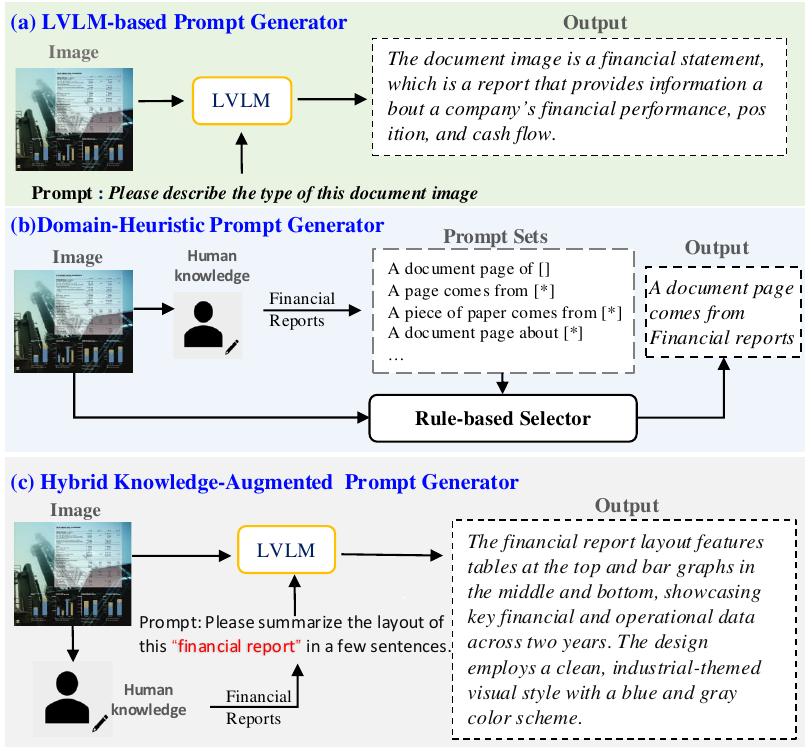}
	\caption{Framework of Prompt Generator.}%(\textbf{a}) The fine-tuning pipeline is used without prompt. (\textbf{b}) Our PromptDLA 
	\label{Prompt Generator}
\end{figure} 

\section{Methodology}
Document layout analysis significantly depends on the data domain. {A model that accurately recognizes domain-specific characteristics can adapt its outputs based on tailored prompts, efficiently handling variations across diverse document domains.}

To facilitate this, we introduce a domain-aware prompter that enables the model to identify the domain of the input data. As illustrated in Fig.~\ref{framework}, our model is composed of four main components: an Image Embedding Module $\mathcal{F}_{patch}$, a Domain-aware Prompter $\mathcal{F}_{prompter}$, a Prompted Transformer Encoder $\mathcal{F}_{encoder}$, and a Detection Head $\mathcal{F}_{detect}$.

Given an input document image $\bm{x} \in \mathbb{R}^{C \times H_{in} \times W_{in}}$, the Image Embedding Module extracts patch embeddings as visual tokens:
\begin{equation}
    \bm{v}_1, \bm{v}_2, ..., \bm{v}_M = \mathcal{F}_{patch}(\bm{x})
    \label{eq:patch_embedding}
\end{equation}
where $\{\bm{v}_i\}_{i=1}^M$ represents the sequence of visual tokens. Currently, the Domain-Aware Prompter generates a domain-specific prompt embedding $\bm{p}_v$:
\begin{equation}
    \bm{p}_v = \mathcal{F}_{prompter}(\bm{x}, d) % Modified to potentially include domain info 'd' or process 'x'
    \label{eq:prompt_embedding}
\end{equation}
where $d$ represents descriptive knowledge as explicit domain information, provided by either a human expert or an LLM, as detailed in Fig.~\ref{Prompt Generator}. Both the visual tokens and the prompt embedding are then processed by the Prompted Transformer Encoder:
\begin{equation}
    \bm{f}_1, \bm{f}_2, ..., \bm{f}_L = \mathcal{F}_{encoder}(\bm{v}_1, \bm{v}_2, ..., \bm{v}_M, \bm{p}_v)
    \label{eq:encoder_output}
\end{equation}
yielding multilevel feature maps $\{\bm{f}_l\}_{l=1}^L$.Finally, a subset of these feature maps is passed to the Detection Head. For instance, when using a transformer-based backbone, features from the 4th, 6th, 8th, and 12th layers are selected to predict the layout structure $\hat{\bm{y}}$:
\begin{equation}
    \hat{\bm{y}} = \mathcal{F}_{detect}(\bm{f}_4, \bm{f}_6, \bm{f}_8, \bm{f}_{12})
    \label{eq:detection_output}
\end{equation}

where $\hat{\bm{y}}$ denotes the predicted refined bounding box $\hat{\bm{b}}$ and a class label $\hat{l}$.

\subsection{Image Embedding}
We follow the patch embedding approach used in ViT~\cite{dosovitskiy2020image} for image embedding. Specifically, the document image $\bm{x}$ is resized to $H \times W$, represented as $\bm{x} \in \mathbb{R}^{C \times H \times W}$, where $C$, $H$, and $W$ denote the channel, height, and width, respectively. The resized image is then divided into a sequence of non-overlapping patches of size $P \times P$. Each patch is linearly projected into a $D$-dimensional vector via $\mathcal{F}_{\text{patch}}$, as defined in Eq.~\ref{eq:patch_embedding}, resulting in a flattened sequence:
\begin{equation}
    \bm{v} = [\bm{v}_1, \bm{v}_2, \dots, \bm{v}_i, \dots, \bm{v}_M], \quad\text{where}\quad \bm{v}_i \in \mathbb{R}^{D}.
\end{equation}
Here, the sequence length $M$ is determined as $M = HW / P^2$. Finally, learnable 1D positional embeddings are added to each vector to encode spatial information.

\subsection{Domain-Aware Prompter}
As shown in Fig.~\ref{framework}, the Domain-aware Prompter $\mathcal{F}_{\text{prompter}}$ consists of two components: a Text Encoder $\mathcal{F}_{t}$ and a Prompt Generator $\mathcal{F}_{g}$.

\subsubsection{Prompt Generator $\mathcal{F}_{g}$ }
\revise{The Prompt Generator $\mathcal{F}_{g}$ is designed to dynamically produce natural language prompts pertinent to the document's domain. As illustrated in Fig.~\ref{Prompt Generator}, $\mathcal{F}_{g}$ supports \revise{three} distinct operational modes, ranging from direct utilization of curated human knowledge to automated generation using large models.}:

\begin{itemize}

    \item \revise{\textbf{ LVLM-based Prompt Generator:} This mode leverages the generative capabilities of LVLMs, such as variants based on LLaMA-Adapter \cite{zhang2024llama}. The input consists of the document image $\bm{x}$ paired with a general instruction (e.g., ``Please describe the type of this document'' or ``Describe the primary use of this document''). The LVLM analyzes the image content and generates a textual description capturing the inferred document domain and characteristics. This approach offers high automation and zero-shot potential but incurs significant computational overhead.}

    \item \revise{ \textbf{Domain-Heuristic Prompt Generator:} This mode relies on curated human knowledge. It utilizes predefined `Prompt Sets` containing various sentence templates designed to encapsulate domain information. Given an explicit domain class (e.g., 'invoice', 'scientific paper'), relevant templates are selected. We employ a rule-based selector, potentially augmented by CLIP's  zero-shot classification capability, to refine template selection. For instance, multiple templates can be instantiated with specific document type names (e.g., from DocLayNet ), embedded using CLIP's text encoder, and evaluated for their zero-shot classification accuracy on a relevant task. Templates yielding top-k performance are retained.}{This approach ensures that the generated prompts are interpretable and consistent with explicit rules and domain labels.}

    \item \revise{ \textbf{Hybrid Knowledge-Augmented Prompt Generator:} This hybrid approach combines aspects previous two. It uses an LVLM but guides its generation with more specific, human-provided knowledge compared to the general instructions. For example, instead of asking the LVLM to infer the type, a prompt like ``Please describe the typical layout elements found in a financial report document'' is provided, potentially along with the image $\bm{x}$ or just the domain label $d$. This allows the LVLM to generate more precise and contextually relevant descriptions tailored to a known document type, balancing automation with targeted knowledge injection.}

\end{itemize}

\begin{figure}[t]
	\centering
	\includegraphics[width=0.8\columnwidth]{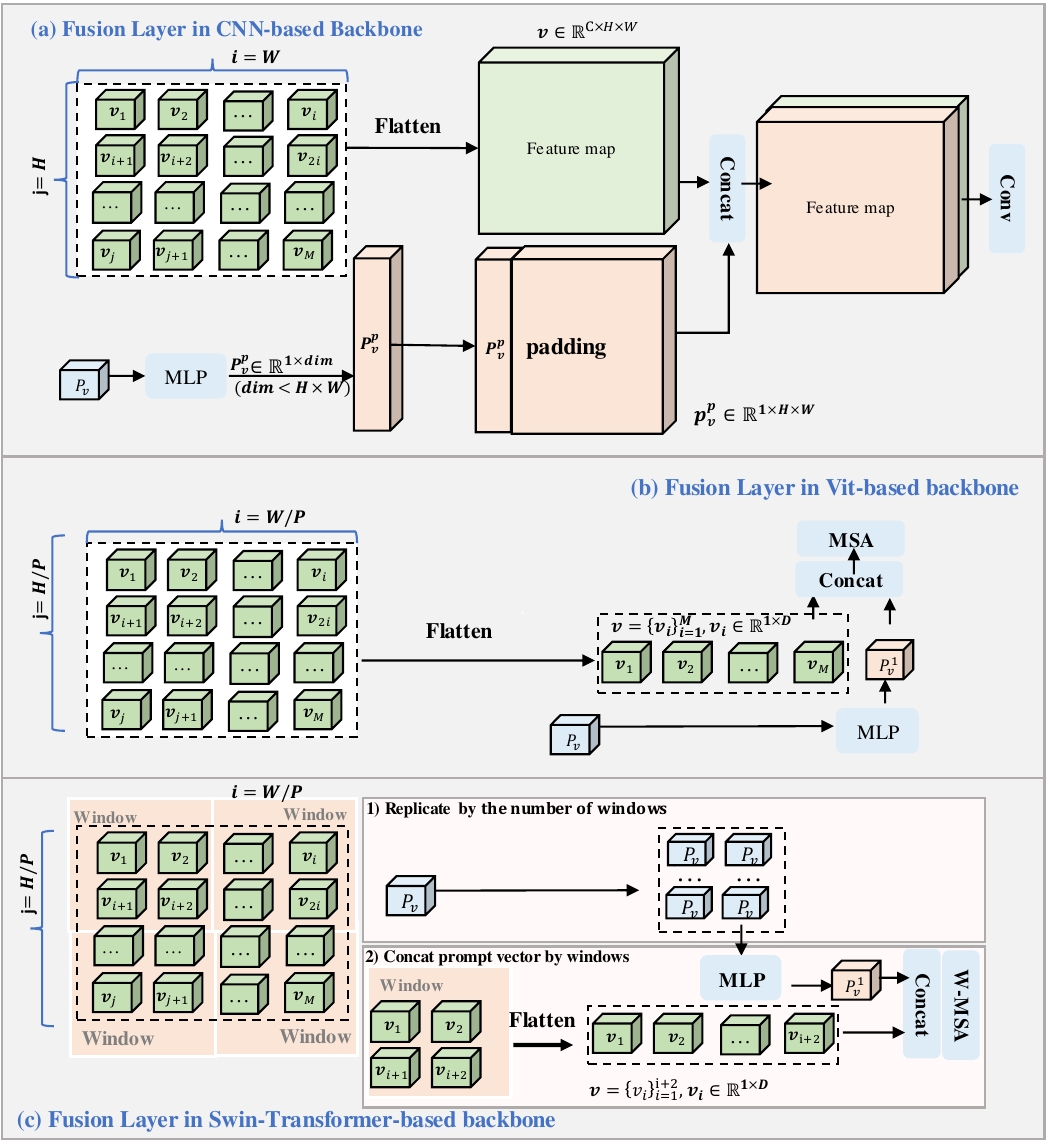}
	\caption{Framework of Fusion Layer.}%(\textbf{a}) The fine-tuning pipeline is used without prompt. (\textbf{b}) Our PromptDLA 
	\label{framefusion}
\end{figure}

\begin{algorithm}[htb]\scriptsize
	\caption{\label{algorithm:CADNDA}The PromptDLA Algorithm. }
	\begin{algorithmic}[1]
		\REQUIRE $D{(x^{(n)},y^{(n)})}^N_{n=1}$, $y$ consists of target bounding boxes $\hat{b}$ and class labels $l$;
		\ENSURE Prediction of layout
		\STATE Initialize \quad $\mathcal{F}_{t} \gets \Theta_{llm}$ or $\Theta_{lvlm}, \enspace \mathcal{F}_{e} \gets \Theta_{pretrain\_e}$ 
		\STATE Prompt Generator $\mathcal{F}_{g}$
		\WHILE{t $\leq$ max iteration}
			\STATE $(v_{1}^0,v_{2}^0,...,v_{m}^0) \gets \mathcal{F}_{\text{patch}}(x)$ 
			\STATE $prompt \gets \mathcal{F}_{g}(\bm{x})$ 
			\STATE $p_v \gets \mathcal{F}_{t}(prompt)$ 
			\FOR{i=1; i $\leq$ layer nums}
				\STATE $f_i \gets (v_1^i,v_2^i,...,v_m^i)$ 
				\STATE $in_i \gets \mathcal{F}_{fuse} (f_i; \mathcal{F}_{mlp}^i(p_v))$ 
				\STATE $(v_1^{i+1},v_2^{i+1},...,v_m^{i+1};p)  \gets \mathcal{F}_{e}^i(in_i)$ 
			\ENDFOR
			\STATE $f \gets \mathcal{F}_{fpn}(f_4, f_6, f_8, f_{12})$ 
            \STATE $Loss \gets R_{loc}(\mathcal{F}_{detect}(f,b),\hat{b}) + \lambda R_{cls}(\mathcal{F}_{detect}(f),l)$ 
			\STATE freeze \enspace $\mathcal{F}_{t}$ 
			\STATE update \enspace $\mathcal{F}_{fuse}$, $\mathcal{F}_{e}$, $\mathcal{F}_{detect}$
            \STATE $t \gets t+1$
		\ENDWHILE
	\end{algorithmic} 
 \label{algri}
\end{algorithm}

\subsubsection{Text Encoder ($\mathcal{F}_{t}$)}
The Text Encoder $\mathcal{F}_{t}$ converts a natural language prompt describing the document domain into a fixed-dimensional embedding that provides domain-specific guidance to the layout analysis model. This Encoder is instantiated using a powerful pre-trained language model, either a text encoder from a vision-language model ($\theta_{\text{lvlm}}$, e.g., CLIP, BLIP2) or a large language model ($\theta_{\text{llm}}$, e.g., LLaMA~\cite{zhang2024llama}). To preserve the rich semantic knowledge acquired during pre-training, the $\mathcal{F}_{t}$ weights are frozen during training. The resulting embedding, denoted as $\bm{p}_v \in \mathbb{R}^{D_{\text{prompt}}}$, serves as the domain-aware input for subsequent modules.

\subsection{Prompted Transformer Encoder} 
The Prompted Transformer Encoder $\mathcal{F}_{encoder}$ integrates the visual tokens $\{\bm{v}_i\}_{i=1}^M$ with the domain prompt embedding $\bm{p}_v$. It typically comprises a standard Transformer Encoder backbone and dedicated Fusion Layers ($\mathcal{F}_{fuse}$) responsible for injecting the prompt information. As shown in Fig.~\ref{framefusion}, We explore compatibility with CNN, ViT and Swin Transformer \cite{liu2021swin} architectures, requiring slightly different fusion strategies:
\begin{itemize}
%TODO Fgiure 5(a) 删除 pv1 
  \item \textbf{CNN-based Encoder :} \revise{As illustrated in Fig.~\ref{framefusion}(a), for a CNN backbone (e.g., ResNet-50 \cite{he2016deep}), the prompt embedding $\bm{p}_v$ is first projected to a target dimension (e.g., 512 or 768) using an MLP layer. This projected prompt vector is then spatially padded to match the height $H$ and width $W$ of the feature map $\bm{F}^{(i)} $ at a specific layer or stage $i$. The resulting projection tensor $p_v^p$, now having dimensions $1 \times H \times W $, is concatenated channel-wise with the feature map $\bm{F}^{(i)} \in \mathbb{R}^{C \times H \times W} $. In our implementation using ResNet-50~\cite{he2016deep}, this fusion operation is performed at the input of each of the four main residual stages, where the spatially expanded prompt features are concatenated with the stage's input feature map.} 

    \item \textbf{ViT-based Encoder:} As depicted in Fig.~\ref{framefusion}(b), for a ViT backbone, the fusion layer $\mathcal{F}_{fuse}$ typically employs an MLP ($\mathcal{F}_{mlp}$) to project the prompt embedding $\bm{p}_v$ to match the visual token dimension $D$. The projected prompt embeddings, denoted $\bm{p}^1_v$, are then concatenated with the sequence of position-aware visual tokens $\{\tilde{\bm{v}}_i\}_{i=1}^M$ (where $\tilde{\bm{v}}_i = \bm{v}_i + \bm{e}_i$, incorporating patch embedding $\bm{v}_i$ and positional embedding $\bm{e}_i$). This combined sequence is the input to the standard Transformer Encoder layers ($\mathcal{F}_e$).

    \item \textbf{Swin Transformer-based Encoder:} For the hierarchical Swin Transformer backbone, illustrated in Fig.~\ref{framefusion}(c), prompt fusion requires adaptation to its windowed attention and shifting window mechanisms \cite{liu2021swin}. Inspired by approaches like VPT \cite{jia2022visual}, the prompt embedding $\bm{p}_v$ is processed by stage-specific MLPs within the fusion module $\mathcal{F}_{fuse}$ to generate dimension-matched embeddings $\bm{p}^{(1)}_v$ for each stage. Within a stage, $\bm{p}_v$ is typically replicated $N_{win}$ times (where $N_{win}$ is the number of windows in that stage) and combined with the token sequence of each respective window before the windowed self-attention (W-MSA) computation. Appropriate masking or padding is applied to ensure dimensional consistency during attention calculations. The outputs from multiple stages (e.g., corresponding to features at 1/8, 1/16, 1/32 resolutions) are then typically fed into an FPN ($\mathcal{F}_{fpn}$) to generate multi-scale features for the detection head.
\end{itemize}

The output of $\mathcal{F}_{encoder}$ (potentially via $\mathcal{F}_{fpn}$) is a set of feature maps $\bm{f}_1, \bm{f}_2, ..., \bm{f}_L$ capturing both visual content and domain-specific context.

\begin{table*}[t]
	\caption{Comparison with state-of-the-art methods on DocLayNet. Our method uses domain prompts from human knowledge and prompt engineering, with a ViT backbone, CLIP text encoder, and Cascade Mask R-CNN detection head. All baseline results are taken directly from previous papers.}

	\label{tab-comp-sota}
	\resizebox{\textwidth}{!}{
		\begin{tabular}{c|c|ccccccccccc|c}
			
			\toprule
			Method    &Pretraining  & Caption & Footnote &Formula & List-item & Page-footer & Page-header & Picture & Section-header & Text & Table & Title   & mAP \\
			\midrule
			Mask R-CNN~\cite{cheng2023m6doc}  & \ding{55}   & 71.5    & 71.8     & 63.4 & 80.8       & 59.3       & 70.0    & 72.7          & 69.3 & 82.9 & 85.8 & 80.4 & 73.5  \\
			Faster R-CNN~\cite{cheng2023m6doc}  &\ding{55}   & 70.1    & 73.7     & 63.5 & 81.0       & 58.9       & 72.0    & 72.0          & 68.4 & 82.2 & 85.4 & 79.9   & 73.4\\
			YOLOv5~\cite{cheng2023m6doc}   & \ding{55}     & 77.7    & 77.2     & 66.2 & \textbf{86.2} & 61.1       & 67.9    & 77.1          & 74.6 & 86.3 & 88.1 & 82.7  & 76.8 \\
			TransDLANet~\cite{cheng2023m6doc} &\ding{55}  & 68.2 & 74.7 & 61.6 & 81.0 & 54.8 & 68.2 & 68.5 & 69.8 & 82.4 & 83.8 & 81.7 & 72.3 \\
			SwinDocSegmenter~\cite{banerjee2023swindocsegmenter}  &\ding{55}      &83.6 & 64.8 & 62.3 & 82.3 & \textbf{65.1} & 66.4   & \textbf{84.7}& 66.5 & \textbf{87.4}& \textbf{88.2}& 63.3  & 76.9 \\
			\hline

			SelfDocSeg~\cite{banerjee2023swindocsegmenter}  &\ding{51}   & - & - & - & -   & - & -    & - & - & -  & - & -   & 74.3 \\
			LayoutLmV3~\cite{huang2022layoutlmv3}   &\ding{51}  & 73.1   & 77.5    & 69.0 & 79.8      & 61.3     & 61.3   & 74.0          & 69.0 & 86.3 & 85.9 & \textbf{84.4}  & 75.7 \\
			DiT~\cite{li2022dit}  & \ding{51}        & 75.0   & 76.2    & 68.1 & 83.5      & 62.1      & 74.0   & 74.5         & 71.2 & 86.4 & 86.6 & 83.0 & 76.4\\
			\midrule
			PromptDLA(ViT, CLIP, Cascade)  &\ding{51}  & 76.6 & 83.0 & \textbf{72.4}& 84.9 & 63.8 & \textbf{76.9}& 75.2 & 73.8 & 87.1 & 87.9 & 84.1 & 78.7\\
            \revise{PromptDLA(ResNet, CLIP, DETR)}  &\ding{51}  & 91.8 & 84.1 &54.7 & 69.4 & 82.9 & 39.5 & 82.0 & \textbf{87.3}& 70.3 & 83.0 & 91.3 & 77.7\\
            \revise{PromptDLA(SwinTran, CLIP, DETR)}  &\ding{51}  & \textbf{92.5} & \textbf{85.5} & 57.6 & 71.3 & 84.2 & 40.2& 83.1 & 88.5& 73.1 & 83.8  & 92.1 &79.6\\
			\bottomrule
	\end{tabular}}
\end{table*}

\subsection{Detection Head}
% This paper utilizes two types of Detection Heads $\mathcal{F}_{detect}$. One is from two-stage detection frameworks, such as Faster-RCNN \cite{ren2015faster}, Mask-RCNN \cite{he2017mask}, or Cascade-RCNN \cite{cai2018cascade}, and the other is from DETR. Our approach is compatible with both detection heads. For the RCNN-based detection frameworks, the CNN backbone is replaced by a Prompt Transformer Encoder. 

The Detection Head $\mathcal{F}_{detect}$ takes the contextualized features from the encoder and performs the final layout element prediction, including bounding boxes and class labels. Our framework is designed to be compatible with two detection head architectures:

\begin{itemize}
    \item \textbf{RCNN-based Heads:} Standard two-stage detection frameworks like Faster R-CNN \cite{ren2015faster}, Mask R-CNN \cite{he2017mask}, or Cascade R-CNN \cite{cai2018cascade} can be readily employed. In this setup,  the R-CNN detection head operates on the generated feature maps by the prompted transformer encoder, optimizing bounding box regression and classification objectives using standard loss functions. Specifically, the bounding box regression loss $R_{loc}$ aims to minimize the discrepancy between predicted boxes $\bm{b}_i = (b_{ix}, b_{iy}, b_{iw}, b_{ih})$ derived from proposals $\hat{\bm{b}}_i$ and features $\bm{f}_i$, and ground-truth boxes $\hat{\bm{b}}_i$:
    \begin{equation}
        R_{loc} = \sum_{i=1}^N L_{loc}(r(\bm{f}_i, \bm{b}_i), \hat{\bm{b}}_i),
        \label{eq:loss_loc}
    \end{equation}
  \revise{where $r(\cdot)$ is the regression function, $N$ is the number of proposals, and $L_{loc}$ is typically the smooth $L_1$ loss}. For Classification, a classifier $c(\cdot)$ assigns feature map patches to one of the classes,  the classification loss  $R_{cls}$ is defined as:
    \begin{equation}
        R_{cls} = \sum_{i=1}^N L_{cls}(c(\bm{f}_i), \bm{l}_i),
        \label{eq:loss_cls}
    \end{equation}
where $\bm{f}_i$ and  $\bm{l}_i$ denote the $i$-th object feature and class label, respectively. And $L_{cls}$ is usually the cross-entropy loss. The total loss is a weighted sum:
    \begin{equation}
        R_{\text{total}} = R_{loc} + \lambda R_{cls}.
        \label{eq:loss_rcnn_total}
    \end{equation}

        \item \textbf{DETR-based Heads:} Our architecture also supports integration with DETR \cite{carion2020end} and its variants. DETR employs an encoder-decoder structure where a set of learnable object queries interacts with the image features (from $\mathcal{F}_{encoder}$, i.e., $\{\bm{f}_l\}_{l=1}^L$) via cross-attention mechanisms within the decoder. Feed-forward networks (FFNs) then directly predict the class and bounding box coordinates from the updated object queries. DETR is optimized end-to-end using a set-based bipartite matching loss that jointly considers classification and localization costs (e.g., cross-entropy for class, and a combination of $L_1$ and GIoU loss for boxes).
    \end{itemize}

This flexibility allows leveraging the strengths of different detection paradigms within our PromptDLA framework.

\section{Experiments}
This section validates PromptDLA's effectiveness through three sets of experiments. First, we examine the efficacy of PromptDLA compared to the state-of-art methods. Next, we evaluate our approach's generalization ability. Lastly, we perform extensive ablation studies and discussions on the model's design.

\subsection{Experimental Settings}
\noindent\textbf{Datasets.}
We conduct extensive experiments to validate the proposed PromptDLA on $5$ DLA benchmark datasets, including different document types, different languages, and different layer styles, 
\begin{itemize}
	\item 
	\textbf{PubLayNet}~\cite{zhong2019PubLayNet} consists of $5$ typical document layout elements: text, heading, list, graphic, and table. It contains over $364232$ page samples, where the annotations were automatically generated by matching PDFs and XML formats of articles from the PubMed Central Open Access subset.	%Following~\cite{gu2021unidoc,li2022dit,zhong2019publaynet}, we train models on the training split (335,703) and evaluate on the validation split (11,245).
	\item
\textbf{DocLayNet}~\cite{pfitzmann2022DocLayNet} contains 6 document types (Financial Reports, Manuals, Scientific Articles, Laws \& Regulations, Patents, and Government Tenders) with 11 categories of annotations across four languages (with English documents comprising approximately 95\%). The dataset contains about 80,863 manually annotated pages.
	\item 
	\textbf{M6Doc}~\cite{cheng2023m6doc} contains a total of $9,080$ modern document images, which are categorized into $7$ document types (Scientific articles, Textbooks,
Books, Test papers, Magazines,
Newspapers, Notes) with $74$ detailed categories. %scientific article, textbook, test paper, magazine, newspaper, note, and book according to content and layouts. %The dataset includes a total of 237,116 annotated instances.
	\item 
	\textbf{D$^4$LA}~\cite{da2023vision} contains a total of $11092$ document images, which includes $12$ diverse document types (Scientific report, Email, Form, Invoice, Letter, Specification,
News article, Presentation, Resume,
Scientific publication, Budget, Memo) with $27$ detailed categories.
		\item
		\textbf{MLDLA} is a \textbf{M}ulti-\textbf{L}anguage \textbf{DLA} (\textbf{MLDLA}) dataset we constructed to evaluate the model generalization on more different languages. It comprises $175,000$ images, which are manually labeled through a uniform labeling style, including $7$ in different languages, such as Persian, Khmer, Kazakh, Lao, Turkish, Hindi, and Vietnamese.
	%	\textbf{DocBank}~\cite{li2020docbank} includes 500K document pages with fine-grained token-level annotations released by Microsoft.	Moreover, region-level annotations in $13$ layout categories (Abstract, Author, Caption, Equation, Figure, Footer, 	List, Paragraph, Reference, Section, Table, and Title) are proposed for object detection. 	We train models on the training split (400K), and evaluate on the validation split (5K)~\cite{li2020docbank}.	
\end{itemize}

%Since both PubLayNet and DocBank datasets are relatively large, we construct two sub-datasets, PubLayNet2K and DocBank2K,  that sample $2,000$ images for training and $2,000$ images for validation, respectively, to quickly verify the effects of different modules of VGT in the early experiments.  $2,000$ images are sampled from training split and $2,000$ image from validation split, respectively. We train VGT for $10,000$ steps on them. Besides three general benchmark datasets, 

\noindent\textbf{Evaluation Metric.}
Our experiments are evaluated using the category-wise and overall mean average precision (mAP) @IOU[0.50:0.95] of bounding boxes following the literature~\cite{cai2018cascade}. This curve describes the relationship between precision and recall and is the most widely used evaluation metric for document layout analysis.
% We use the evaluation metric of the category-wise and overall mean average precision (mAP) @IOU[0.50:0.95] of bounding boxes.

\noindent\textbf{Implementation Details.}
We train our model on $8$ 3090 GPUs with a batch size 16 using a cosine learning rate schedule and a warm-up strategy with a 0.01 warm-up factor. We set the basic learning rate to 2e-4. Additionally, we adopt the AdamW optimizer. Our study uses DiT~\cite{li2022dit} pre-trained weights and the Cascade-RCNN~\cite{cai2018cascade} detection head as our baseline method. The detailed training process is shown in Algorithm~\ref{algorithm:CADNDA}. %Section 4.5 of our study describes the interchangeable components of our method in the ablation study.

\subsection{Generalization Ability}
%We validate the generalization ability of PromptDLA towards different document domains and inconsistent labeling style as shown in Table~\ref{tab:pretrain} and Table~\ref{exp-incosistent-label},respectively. %and Table~\ref{noisy-exp}.

\noindent\textbf{Generalization on Different Document Domains.} 
We evaluate the performance of PromptDLA on different DLA datasets with more diverse document domains in Table~\ref{tab:pretrain}. The results show that the promptDLA can get consistent improvements on datasets with other domains, such as DocLayNet with $6$ domains (2.3\% over DiT),  M6Doc with $7$ domains( 2.0\% over DiT), and D4LA (1.4\% over DiT), validating the generalization of PromptDLA on different document types.

%\begin{table}[t]\small
%	%	\setlength{\tabcolsep}{1.5pt}
%	%	\renewcommand{\arraystretch}{1}
%	\caption{Experiments on Different Document Types. }
%	\label{tab:doctype}
%\begin{tabular}{c|c|cc}
%	\hline
%	 Dataset & \#Document Type & DiT-Base & PromptDLA \\
%	\hline 
%	 DocLayNet & 6 & 76.4 & \textbf{78.7}(\blue{+2.3}) \\
%	 M6Doc & 7 & 67.2 & \textbf{69.2}(\blue{+2.0}) \\
%     D4LA & 12 & 67.7 & \textbf{68.7}(\blue{+1.0}) \\
%	\hline
%\end{tabular}
%\end{table}

\noindent\textbf{Generalization on Multi-Language Datasets.}
We investigate the effects of domain-aware prompts by using different language types as domain information based on the MLDLA dataset.
Firstly, we validate that CLIP can provide prior knowledge even for documents in different languages, including minority languages, through a zero-shot document classification task. More specifically, we insert the document's language into a prompt template as text and utilize CLIP to identify similarities between it and the associated image. Our experiments indicate that CLIP achieves a zero-shot classification accuracy of \textbf{47.53\%} across seven languages in MLDLA. 
Furthermore, we apply PromptDLA to MLDLA and present a comparison in Table~\ref{exp-mlt}. Following the method described in Section 3.2 of the paper to construct prompt sets.
Our approach improves precision by +1.0 compared to the DiT model without a prompt. DiT without a prompt achieves higher mAPs in categories like "List," which are less domain-relevant. We think that it may lack sufficient domain-specific information for accurate detection.  On the contrary, our PromptDLA significantly improves other domain-specific details, such as "Fig." (from 54.5\% to 57.3\%) and "Table" (from 76.1\% to 77.9\%).

% \begin{figure}[t]
% 	\centering
% 	\includegraphics[width=0.8\columnwidth]{images/supply-MLDLA.pdf}
% 	\caption{Domain information and prompt sets for MLDLA.}%(\textbf{a}) The fine-tuning pipeline is used without prompt. (\textbf{b}) Our PromptDLA 
% 	\label{supply-mldla}
% \end{figure} 

\begin{table}[!htbp]
	\caption{Experiments on a real multi-language dataset.}
	\begin{tabular}{p{1.4cm}p{0.7cm}p{0.7cm}p{0.7cm}p{0.7cm}p{0.7cm}p{0.7cm}<{\centering}}
		\hline
		Method    & Text  & Title & Figure & List  & Table & mAP  \\
		\hline
		DiT       & 77.9 & 58.6  & 54.5  &\textbf{75.6} & 76.1  & 68.5\\
		PromptDLA   & \textbf{78.7} & \textbf{59.1} & \textbf{57.3}  & 74.5  & \textbf{77.9}  & \textbf{69.5}\\
		$\Delta$   &$\blue{+0.8}$ &$\blue{+0.5}$ &$\blue{+2.8}$ & $\red{-1.1}$ & $\blue{+1.8}$  &$\blue{+1.0}$\\ 
		\hline
	\end{tabular}
	\label{exp-mlt}
\end{table}

\begin{figure}[h]
	\centering
	\includegraphics[width=0.73\columnwidth]{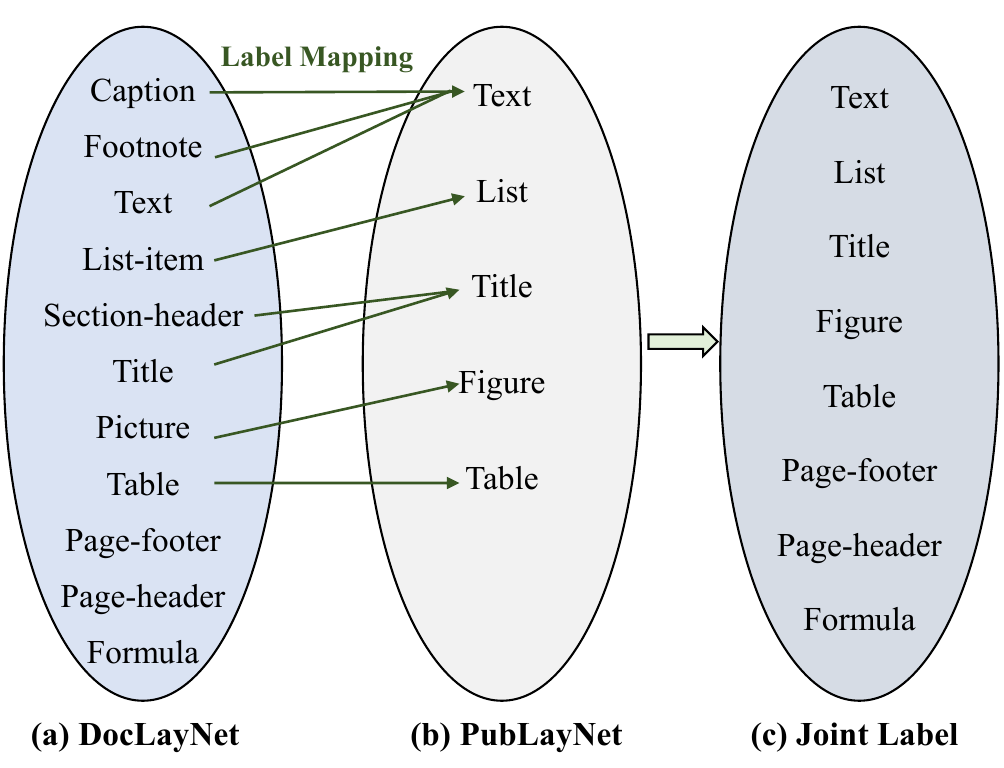}
	\caption{Inconsistent labeling relationship between DocLayNet and PubLayNet.}%, where blue, yellow, and red lines mean that labels are the same, containment, and different, respectively.}
	\label{pub-doc}
\end{figure}

\noindent\textbf{Generalization on Inconsistent Labeling Style}. 
We investigate the impact of inconsistent labeling styles. We perform experiments to jointly train different datasets to enhance the model's performance in practical applications. However, we encounter challenges due to the inconsistent labeling styles observed in public Document Layout Analysis (DLA) datasets, particularly between DocLayNet and PubLayNet.
As depicted in Fig.~\ref{pub-doc}, the labeling styles of PubLayNet and DocLayNet exhibit notable differences. While PubLayNet's image and table align with DocLayNet's counterparts, PubLayNet's text corresponds to the set of DocLayNet's caption, footnote, and text. Similarly, PubLayNet's title matches the set of DocLayNet's title and section header. Notably, PubLayNet's list and DocLayNet's list items differ, with PubLayNet integrating multiple list items as a whole list, while DocLayNet labels each list item separately. Additionally, PubLayNet omits page-footer, page header, and formula, whereas DocLayNet includes these elements. To address these differences, we perform label mapping, which aligns DocLayNet labels with PubLayNet and retains the page footer, page header, and formula.
As shown in Table~\ref{exp-incosistent-label}, joint training of the datasets, even after label mapping, does not improve performance on DocLayNet. Instead, there is a decrease in performance attributed to annotation conflicts. To overcome this issue, we introduce domain prompts, observing a consistency improvement on both DocLayNet and PubLayNet. This confirms the model's adaptive learning capability to handle conflicts and effectively learn models tailored to the target domain. Notably, our method enhances mAP from 76.0 to 77.1 for DocLayNet and from 94.8 to 94.9 for PubLayNet.

\begin{table}[h]
	\centering
	\caption{Experiments on inconsistent labeling styles.}
	\renewcommand{\arraystretch}{1}
	\resizebox{\linewidth}{!}{
		\begin{tabular}{c|ccc|ccc}
			\hline
			& \multicolumn{3}{c}{DocLayNet} & \multicolumn{3}{c}{PubLayNet} \\
			& Baselines  & Joint      & PromptDLA                            & Baselines & Joint     & PromptDLA      \\
			\hline
			Text          & 86.9  & 87.0(\blue{+0.1})   & \textbf{87.5}(\blue{+0.6})     & 94.4   & 94.5(\blue{+0.1})  & \textbf{94.5}(\blue{+0.1})      \\		Title         & 71.2  & 72.1(\blue{+0.9})   & \textbf{73.7}(\blue{+2.5})     & 88.9   & 89.4(\blue{+0.5})  & \textbf{89.5}(\blue{+0.6})      \\
			List          & 84.0  & 83.5(\red{-0.5})    & \textbf{84.4}(\blue{+0.4})     & 94.8   & 95.5(\blue{+0.7})  & \textbf{95.6}(\blue{+0.8})       \\
			Table         & 87.1  & 87.2(\blue{+0.1})   & \textbf{87.7}(\blue{+0.6})     & 97.6   & 97.8(\blue{+0.2}) & 97.8(\blue{+0.2})       \\
			Figure        & 77.2  & 75.7(\red{-1.5})    & \textbf{76.3}(\red{-0.9})      & 96.9   & 96.9(\blue{+0.0})  & \textbf{97.0}(\blue{+0.1})       \\
			Formula       & 68.4  & 69.1(\blue{+0.7})   & 68.9(\blue{+0.5})              & -  & -  & -         \\
			Page-footer   & 64.0  & 61.7(\red{-2.3})    & \textbf{64.1} (\blue{+0.1})    & -  & -  & -        \\
			Page-header   & 74.0  & 72.1(\red{-1.9})    & \textbf{74.3} (\blue{+0.3})    & -  &-   & -        \\
			mAP           & 76.6  & 76.0(\red{-0.6})    & \textbf{77.1} (\blue{+0.5})    & 94.5   & 94.8(\blue{+0.3})  & \textbf{94.9}(\blue{+0.4})  \\
			\hline
	\end{tabular}}
	\label{exp-incosistent-label}
\end{table}

\noindent\textbf{Generalization on Out of Distribution Prompt.} We explored the performance of PromptDLA in out-of-distribution (OOD) scenarios by splitting the DocLayNet dataset by document category, using 'manuals' as the test set, and training on the remaining categories of documents. As shown in Table~\ref{ood_manuals}, the prompt works effectively in OOD situations.

\begin{table}[h]
	\caption{OOD result on Manuals from DocLayNet}
	\centering
	\begin{tabular}{c|ccc}
		\hline
		Tag&	Method     & mAP   & $\Delta$                \\
		\hline
		(a)&	DiT   & 62.68 & \\
		(b)&	DiT + Human knowledge         &   64.23 & +1.55\\
		(c)&	DiT + LVLM 		& 63.93 & +1.25\\
		%		PromptDLA  & 78.69 &                         \\
		%		Shallow Prompted Encoder  & 78.41 & -0.28    \\
		%		Share MLP Layer & 78.44 & -0.25             \\
		\hline
	\end{tabular}
	\label{ood_manuals}
\end{table}

% \noindent\textbf{Generalization on Different Framework}
% Our prompts can be embedded into various backbones, including ViT, Swin Transformer, and ResNet50, all demonstrating effectiveness. The results are presented in Table~\ref{exp-frame}.

\noindent\textbf{\revise{Generalization Across Different Backbone Architectures.} } 
\revise{ To evaluate the adaptability and robustness of our proposed PromptDLA framework,  we integrated and tested it with diverse backbone architectures commonly used in vision tasks. Specifically, we assessed its performance with a standard Vision Transformer (ViT-Base), a hierarchical Vision Transformer (Swin-Transformer Base), and a widely adopted Convolutional Neural Network (ResNet-50).
For the Transformer-based models (ViT and Swin-Transformer), the domain-aware prompts were incorporated as detailed in *Figure*.
For the CNN-based ResNet-50, the prompt embeddings were integrated by concatenating them with the pooled features before the final classifier.
The performance, measured by mean Average Precision (mAP), was evaluated on the DocLayNet dataset.} 
\revise{Table~\ref{tab:exp-backbone-comparison} compares these backbones with and without the PromptDLA module.
PromptDLA consistently enhances the performance across all tested architectures. Notably, it improved +2.3 mAP points for ViT-Base, +1.0 mAP points for Swin-Base, and +0.7 mAP points for ResNet-50 compared to their respective baselines. These consistent gains underscore the versatility of PromptDLA, demonstrating that its effectiveness is not confined to a specific architectural paradigm and that it successfully leverages domain cues to benefit both Transformer and CNN models in document layout analysis.}

\begin{table}[h]
	\caption{\revise{Performance comparison of different backbone architectures with and without PromptDLA on the DocLayNet dataset. $\Delta$ indicates the absolute mAP improvement achieved by adding PromptDLA compared to the respective baseline model}}
	\centering
	\begin{tabular}{c|ccc}
		\hline
		Tag&	Method     & mAP   & $\Delta$                \\
		\hline
		(a)&	ViT   & 76.4 & \\
		(b)&	ViT + PromptDLA         &   78.7 & +2.3\\
  \hline
		(c)&	Swin 		& 78.7 & -0.12\\
        (d)&	Swin + PromptDLA		& 79.7 & +1.0\\
        \hline
        (e)&	ResNet50 (DETR) 		& 77.0 & \\
        (f)&	ResNet50 + PromptDLA	& 77.7 & +0.7\\
		%		PromptDLA  & 78.69 &                         \\
		%		Shallow Prompted Encoder  & 78.41 & -0.28    \\
		%		Share MLP Layer & 78.44 & -0.25             \\
		\hline
	\end{tabular}
    \label{tab:exp-backbone-comparison}
\end{table}

\textbf{Generalization Across Different Detection Heads.}
We further evaluate the generalizability of PromptDLA by integrating it with two representative detection heads: Faster R-CNN and DETR, both using a ResNet-50 backbone.
As shown in Table~\ref{tab:exp-head-comparison}, PromptDLA improves mAP by +1.6 on Faster R-CNN and +0.7 on DETR, demonstrating its compatibility with both traditional and Transformer-based detectors.

\begin{table}[h]
	\caption{Effect of PromptDLA on different detection heads. $\Delta$ denotes the absolute mAP improvement. We re-implemented baseline models for fair comparison.}
	\centering
	\begin{tabular}{c|ccc}
		\hline
		Tag &	Method             & mAP   & $\Delta$ \\
		\hline
		(a) &	Faster R-CNN (R50)       & 66.8  &         \\
		(b) &	Faster R-CNN + PromptDLA & 68.4  & +1.6     \\
		\hline
		(c) &	DETR (R50)               & 77.0 &         \\
		(d) &	DETR + PromptDLA    & 77.7 & +0.7    \\
		\hline
	\end{tabular}
    \label{tab:exp-head-comparison}
\end{table}

% --- Revised Table ---

% Use [htbp] for better float placement flexibility than just [h]
% \begin{table}[htbp]
% \centering
% % Add a more descriptive caption: What metric? What dataset? What does Delta mean?
% \caption{Performance comparison (mAP \%) of different backbone architectures with and without PromptDLA on the [**Specify Dataset Name**] dataset. $\Delta$ indicates the absolute mAP improvement achieved by adding PromptDLA compared to the respective baseline model.}
% % Use a more descriptive label, common convention is tab:
% \label{tab:exp-backbone-comparison}
% % Use booktabs style for better horizontal lines and spacing. Adjust column alignment (l=left, c=center, r=right) as needed.
% \begin{tabular}{@{}l l c c@{}} % l=left align method names, c=center others. @{} removes extra padding at table edges.
%     \toprule
%     Tag & Method             & mAP (\%) & $\Delta$ \\ % Add (%) to mAP for clarity
%     \midrule
%     (a) & ViT-Base           & 76.4     & --       \\ % Add -Base suffix if it was the base model. Use -- for baseline delta.
%     (b) & ViT-Base + PromptDLA & 78.7     & +2.3     \\
%     \midrule % Separate different backbone groups
%     (c) & Swin-Base          & 78.7     & --       \\ % Corrected: Baseline has no delta relative to itself
%     (d) & Swin-Base + PromptDLA& 79.7     & +1.0     \\
%     \midrule
%     (e) & ResNet50           & 77.0     & --       \\ % Corrected: Baseline has no delta
%     (f) & ResNet50 + PromptDLA & 77.7     & +0.7     \\
%     \bottomrule
% \end{tabular}
% \end{table}

\subsection{Comparison with State-of-the-arts}
We compare the proposed PromptDLA with
(1) the DLA frameworks without pretraining: Faster-RCNN~\cite{ren2015faster}, Mask-RCNN~\cite{he2017mask}, YOLOV5~\cite{zhu2021tph}, SwinDocSegmenter~\cite{banerjee2023swindocsegmenter}, and TransDLANet~\cite{cheng2023m6doc} and (2) the DLA framework with different pretraining models: DiT~\cite{li2022dit}, LayoutLMv3~\cite{huang2022layoutlmv3}, and SelfDocSeg~\cite{banerjee2023swindocsegmenter}. 
%This section evaluates the model's performance on Doclaynet\cite{pfitzmann2022DocLayNet}. The DocLayNet dataset includes document images from six disciplines: financial reports, manuals, laws and regulations, government tenders, patents, and scientific articles. Our approach utilizes document type as a domain class for the domain prompt.
We evaluate PromptDLA on the challenging DocLayNet dataset~\cite{pfitzmann2022DocLayNet}, which provides diverse document types with significant domain variations. In our experimental setup, we systematically use the document type as the domain prompts, enabling our model to adapt to specific document characteristics. 
Table~\ref{tab-comp-sota} presents that our PromptDLA outperforms the DLA framework with and without pretraining methods.
We can observe that the promptDLA outperforms state-of-the-art SwinDocSegmenter~\cite{banerjee2023swindocsegmenter} with 1.8\% mAP. Although SwinDoc gets better mAPs in a few rows like "Table" and "Picture," we think these discernible categories are less relevant to the domain. Nevertheless, our PromptDLA exhibits substantial improvement over other domain-related detail categories, such as "Footnote" (from 64.8 to 83.0) and "Section-Header" (from 66.4 to 76.9).

% Specifically, the accuracy of PromptDLA improves by +1.84 compared to SwinDocSegmenter, by +1.89 compared to YOLOV5x6, by +2.26 compared to DiT, and by +2.96 compared to LayoutLMv3.

\subsection{Performance on Different Pretrained Models}
We assess the performance of PromptDLA upon different pre-trained DLA models and datasets, validating that PromptDLA is easily plugged to enhance different DLA frameworks. As shown in Table~\ref{tab:pretrain}, our method can be applied to different pre-trained frameworks, including LayoutlmV3 and DiT. The performance of the PromptDLA is a further improvement on the pre-trained model. The stronger the performance of the pre-trained model, the better our method performed based on it. 
Excitingly, PromptDLA outperforms state-of-the-art models such as SwinDocSegmenter~\cite{banerjee2023swindocsegmenter}, TransDLA~\cite{cheng2023m6doc}, and VGT~\cite{da2023vision} by 1.8\%, 5.4\%, and 0.3\% on DocLayNet, M6Doc, and D$^4$LA, respectively. Notably, compared to tailored models like SwinDocSegmenter and VGT, DiT with PromptDLA stands out for its simplicity and effectiveness.

% The promptDLA can get consistent improvement on LayoutLMv3 and DiT, validating the effectiveness of the promptDLA. 

\begin{table}[t]\small
	\caption{Performance of the PromptDLA with different pre-trained models on DocLayNet, M6Doc, and D$^4$LA, and comparison with the current state-of-the-art method}
    \centering
	\label{tab:pretrain}
	\resizebox{0.9\linewidth}{!}{
		\begin{tabular}{cccc}
			\hline 
			\multirow{2}{*}{ Model } & \multicolumn{3}{c}{ mAP@IOU[0.50:0.95]}  \\
			\cline { 2 - 4 }  & DocLayNet & M6Doc & D$^4$LA  \\
			\hline 
			TransDLA~\cite{cheng2023m6doc}			                & 72.3 		& 63.8   	& - \\
			SwinDocSegmenter~\cite{banerjee2023swindocsegmenter}	& 76.9 		& -   		& - \\
			VGT~\cite{da2023vision}				                    & - 		& -   		& 68.8\\
			
			\hline
			LayoutLMv3 		  & 75.7 				& 60.5 				& 62.6 \\
			\quad  +PromptDLA & 76.4(\blue{+0.7})   & 61.3(\blue{+0.8}) & 63.1(\blue{+0.5}) \\
			\hline
			DiT				  & 76.4 						  & 67.2   					    & 67.7\\
			\quad +PromptDLA  & \textbf{78.7}(\blue{+2.3})    & \textbf{69.2}(\blue{+2.0})  & \textbf{69.1}(\blue{+1.4}) \\
			\hline
	\end{tabular}}
\end{table}

\subsection{Different Prompter}
\noindent\textbf{Different pre-trained text encoder for PromptDLA}
\begin{figure}[t]
	\centering
	\includegraphics[width=0.85\columnwidth]{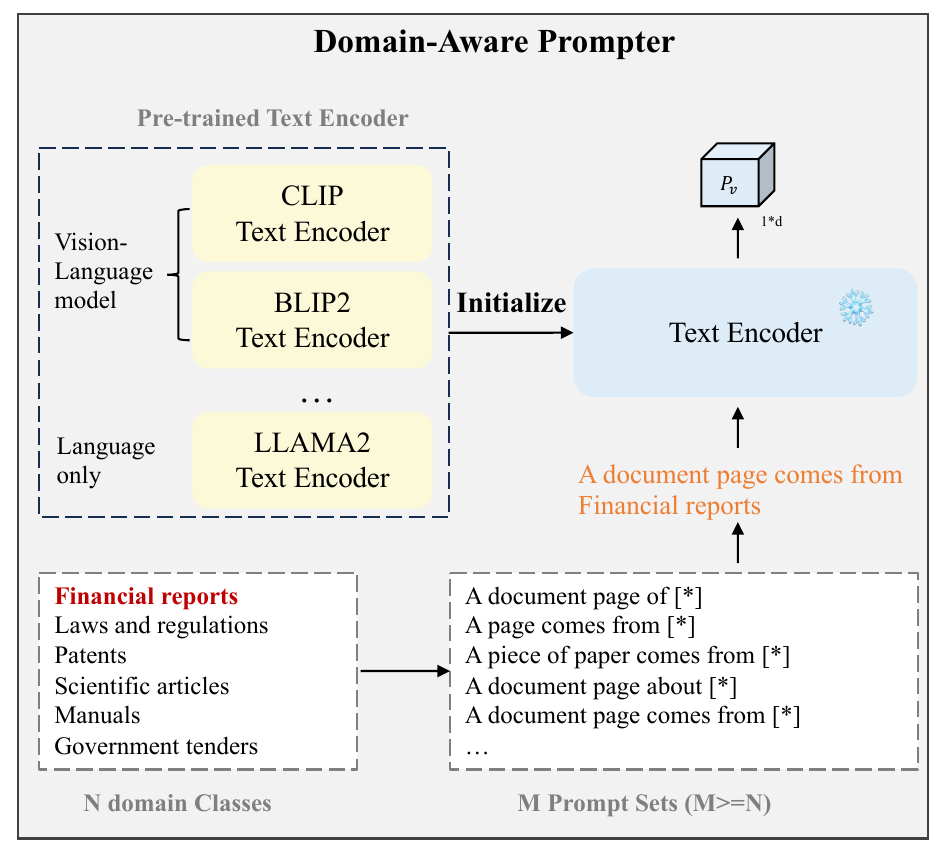}
	\caption{Different pre-trained text encoder for PromptDLA.}%(\textbf{a}) The fine-tuning pipeline is used without prompt. (\textbf{b}) Our PromptDLA 
	\label{supply-other-text-encoder}
\end{figure} 
%In this section, we further explore how much prior knowledge about the document image can be provided by the different large pre-trained models. We test the performance of the text encoder from different large pre-trained models. Pre-trained text encoders can be classified into two categories: those from the vision-language pre-trained model and those from the language-only pre-trained model. We explored the CLIP, BLIP2\cite{li2023blip}, and LLAMA2\cite{touvron2023llama} models. Both CLIP and BLIP2 are vision-language models, while LLAMA2 is only a language 
we delve deeper into the extent of prior knowledge that various large pre-trained models can provide about document images. We evaluate the performance of text encoders derived from different large pre-trained models, which fall into two main categories: those originating from vision-language pre-trained models and those from language-only pre-trained models. Specifically, we explore the capabilities of CLIP~\cite{radford2021learning}, BLIP2\cite{li2023blip}, and LLAMA2\cite{touvron2023llama}. Notably, CLIP and BLIP2 are vision-language models, while LLAMA2 exclusively operates as a language model.

As illustrated in Table~\ref{other text encoder}, first, compared to the baseline model DiT without any domain-aware prompt, "w/o Pre-trained Text Encoder" reduces the model's accuracy by 0.1. This suggests that utilizing domain information but randomly initializing it without a pre-trained text encoder doesn't effectively guide the model to differentiate between various domain documents. Conversely, using weights from pre-trained models improves performance for all three models, emphasizing a solid correlation between document images in DLA datasets and their respective domains. It highlights the ability of the Text Encoder from Pre-trained models to provide valuable prior knowledge.

\begin{table}[!htbp]
	\centering
	\caption{Experiments on different Pre-trained model Text Encoders (w/o Pre-trained Text Encoder means random initializing prompt without using a Pre-trained Text Encoder).}
	\resizebox{0.7\linewidth}{!}{
		\begin{tabular}{ccc}
			\hline
			& mAP            &$\Delta$  \\
			\hline
   			DiT~\cite{li2022dit} & 76.4           &          \\
			w/o Pre-trained Text Encoder & 76.3             &\red{$-0.1$}          \\
			CLIP Text Encoder~\cite{radford2021learning}        & 78.7           &\blue{$+2.3$}    \\
			BLIP2 Text Encoder~\cite{li2023blip}             & 79.0           &\blue{$+2.6$}    \\
			LLAMA2 Text Encoder~\cite{touvron2023llama}      & 77.8           &\blue{$+1.4$}    \\
			\hline
	\end{tabular}}
	\label{other text encoder}
\end{table}

Moreover, CLIP and BLIP2 outperform LLAMA2, indicating that the Text Encoder from vision-language pre-trained models is superior to language-only pre-trained models. A vision-language pre-trained model can offer prior knowledge about both the relationship between document images and their corresponding text descriptions and general text representation. In contrast, a language-only pre-trained model only possesses knowledge concerning understanding human language.

Furthermore, BLIP2 outperforms CLIP by 0.3 when using the same prompt, suggesting that BLIP2 can more accurately find the relationship between document images and their text descriptions. Consequently, a visual language pre-training model with superior performance could provide an even greater boost to my approach. In our paper, we uniformly use CLIP to explore the role of other modules, but we can replace it with BLIP2 or other superior visual language pre-training models for more accurate results.

% \noindent\textbf{Human Knowledge vs LVLMs.}
% As shown in Fig.~\ref{Prompt Generator}, \revise{our Prompter comes in three types: one derived from human prior knowledge, another from a Large Vision Language Model, and the final one being a hybrid.} Additionally, we have independently trained a classifier using human prior knowledge to categorize document types for use in real-world applications where the document category is not provided. For the six document categories of DocLayNet, the ViT-base classifier achieves an accuracy of 90\%, and the classifier's error has a very minimal impact on the overall detection performance, as shown in Table~\ref{exp-struc}. The results from the Large Vision Language Model are similar to those from human prior knowledge. The advantage of the Large Vision Language Model is that it does not require predefined document categories, making it more general. However, the drawback is that it consumes more computational resources. Introducing an additional classifier slightly reduces mAP by 0.12, so the impact is minimal. \revise{In addition, we explore using both human knowledge and LVLMs to generate prompts, as shown in Fig.~\ref{Prompt Generator}(c). With human knowledge, the LVLMs can generate more accurate prompts, leading to a 0.33 improvement in results. Furthermore, we explore using two prompts simultaneously, as shown in Table~\ref{exp-struc}(e). With nearly no improvement compared to (d), a better prompt is more effective than using multiple prompts.}

\noindent\textbf{Analysis of Prompt Generation Strategies.}
We investigate different prompt generation strategies as illustrated in Fig.~\ref{Prompt Generator}. As shown in Table~\ref{exp-struc-prompt}, we first compare human knowledge versus LVLM generation approaches. For real-world applications where domain labels are unavailable, human-knowledge prompts require a preceding classification step. We trained a ViT-based domain classifier that achieves 90\% accuracy on DocLayNet. Integrating this classifier (c) results in a negligible mAP drop of only 0.12 compared to using ground-truth labels (a), confirming the approach's practical viability. In contrast, the LVLM-based generator (b) offers greater generality by eliminating the need for predefined categories, albeit at a higher computational cost. Despite this difference, its performance remains nearly identical to the human-knowledge approach.
Second, we examine hybrid prompts and their impact on prompt quality. To combine the strengths of both methods, our hybrid strategy (d) leverages human knowledge to guide the LVLM, as shown in Fig.~\ref{Prompt Generator}(c). This approach performs best, improving the mAP by 0.33 over the baseline, demonstrating that high-level human knowledge can effectively steer the LVLM toward more relevant feature descriptions. Furthermore, we explored concatenating prompts from methods (a) and (b) simultaneously (e). This concatenation resulted in almost no improvement, underscoring that the quality and relevance of a single, well-formed prompt are more critical than prompt quantity.

\begin{table}[h]
	\caption{Analysis of different prompt generation strategies on the DocLayNet dataset. The $\Delta$ column measures the mAP change relative to the baseline Human-Knowledge Prompt (a).}
	\centering
	\begin{tabular}{c|l|cc}
		\hline
		Tag & Method     & mAP   & $\Delta$                \\
		\hline
		(a) & Human-Knowledge Prompt (Baseline)         & 78.69 & - \\
		(b) & LVLM-Generated Prompt                    & 78.68 & -0.01 \\
		(c) & Human-Knowledge Prompt w/ Classifier      & 78.57 & -0.12 \\
        \revise{(d)} & \revise{Hybrid Prompt (Human-Guided LVLM)}      & \revise{79.02} & \revise{+0.33} \\
        \revise{(e)} & \revise{Concatenated Prompts (a) + (b)}      & \revise{78.71} & \revise{+0.02} \\
		\hline
	\end{tabular}
	\label{exp-struc-prompt}
\end{table}

% \begin{table}[h]
% 	\caption{Analysis of Prompt Generation Strategies on DocLayNet}
% 	\centering
% 	\begin{tabular}{c|ccc}
% 		\hline
% 		Tag&	Method     & mAP   & $\Delta$                \\
% 		\hline
% 		(a)&	Domain-Heuristic Prompt Generator   & 78.69 & \\
% 		(b)&	LVLM-based Prompt Generator         &   78.68 & -0.01\\
% 		(c)&	+ Extra classifier 		& 78.57 & -0.12\\
%         (d)& Hybrid Knowledge-Augmented Prompt Generator & 79.02 & +0.33\\
%         (e) &(a)+(b) &78.71 &+0.02\\
% 		\hline
% 	\end{tabular}
% 	\label{exp-struc}
% \end{table}

% \begin{table}[h]
% 	\caption{Analysis of Prompt Generation Strategies on DocLayNet}
% 	\centering
% 	\begin{tabular}{c|ccc}
% 		\hline
% 		Tag&	Method     & mAP   & $\Delta$                \\
% 		\hline
% 		(a)&	Domain-Heuristic Prompt Generator   & 78.69 & \\
% 		(b)&	LVLM-based Prompt Generator         &   78.68 & -0.01\\
% 		(c)&	+ Extra classifier 		& 78.57 & -0.12\\
%         \revise{(d)}& \revise{Hybrid Knowledge-Augmented Prompt Generator} & \revise{79.02} & \revise{+0.33}\\
%         \revise{(e)} &\revise{(a)+(b)} &\revise{78.71} &\revise{+0.02}\\
% 		%		PromptDLA  & 78.69 &                         \\
% 		%		Shallow Prompted Encoder  & 78.41 & -0.28    \\
% 		%		Share MLP Layer & 78.44 & -0.25             \\
% 		\hline
% 	\end{tabular}
% 	\label{exp-struc}
% \end{table}

\subsection{\revise{Multi-Modalities}}
\revise{We finally explore whether the text of a document can improve layout analysis accuracy. We use Optical Character Recognition (OCR) to extract the text information from the document and then use the CLIP text encoder to retrieve text tokens. These tokens are concatenated into the PromptDLA backbone. The results, shown in Table~\ref{ocr}, indicate that the text modality does not improve the layout analysis performance.}
\begin{table}[h]
	\caption{\revise{Comparison of Visual-Only vs Textual and Visual Modalities}}
	\centering
	\begin{tabular}{c|ccc}
		\hline
		Tag&	Method     & mAP   & $\Delta$                \\
		\hline
		(a)&	PromptDLA   & 78.69 & \\
		(b)&	PromptDLA + OCR         &   78.53 & -0.16\\
		\hline
	\end{tabular}
	\label{ocr}
\end{table}

\begin{figure}[h]
	\centering
	\includegraphics[width=0.95\columnwidth]{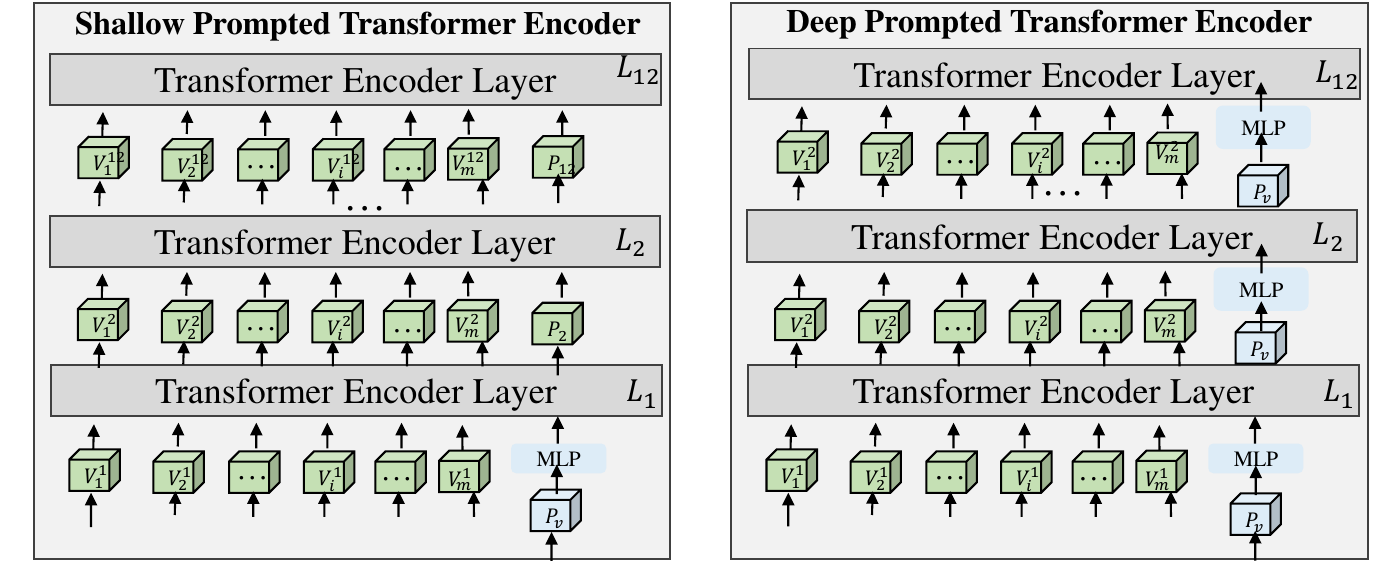}
	\caption{Details of the shallow and deep prompt.}
	\label{deep-shallow}
\end{figure}

\begin{figure*}[h]
	\centering
	\includegraphics[width=0.9\linewidth]{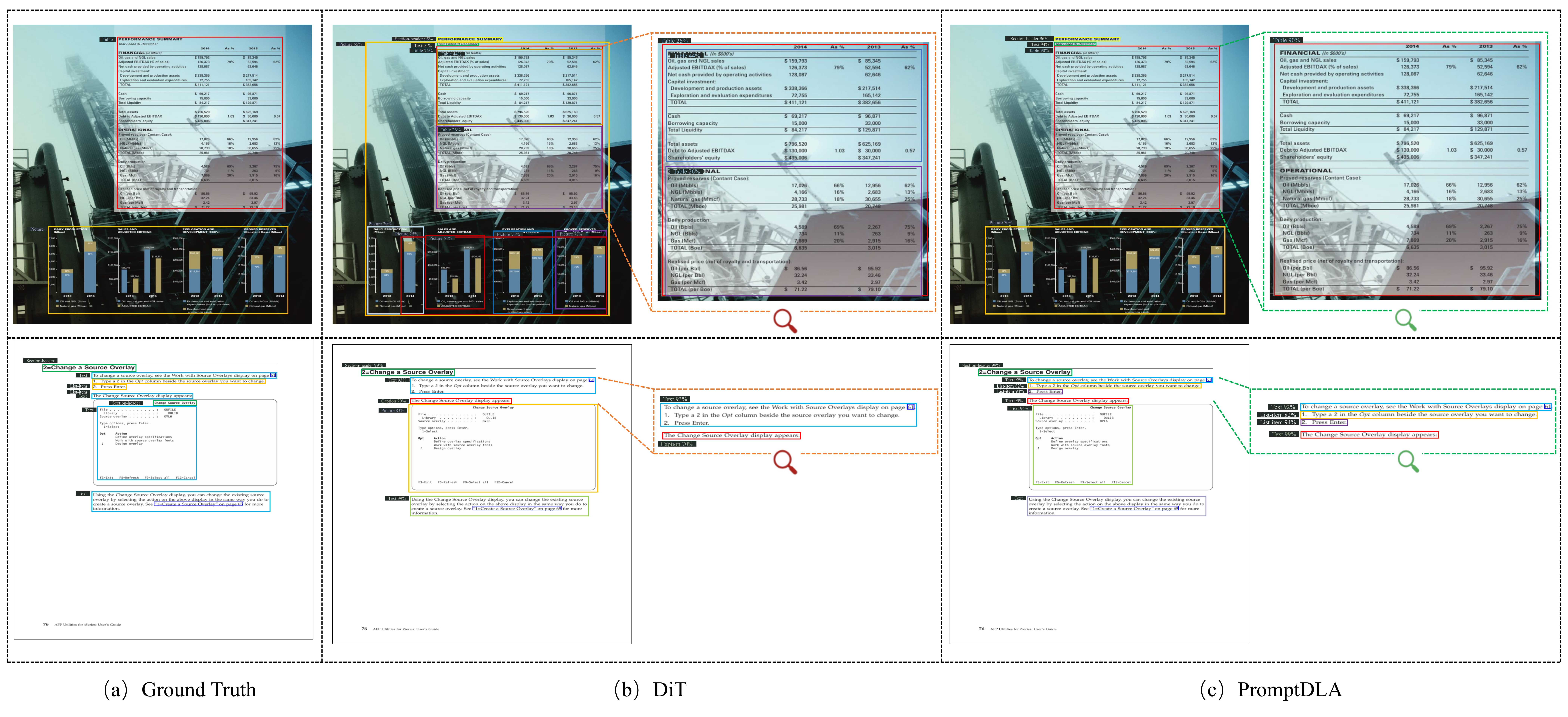}
	\caption{Qualitative comparison between DiT and PromptDLA on Financial Reports (1st row) and Laws (2nd row) domain from DocLayNet. It is best viewed in color and zooming out.}
	\label{vi DocLayNet result}
\end{figure*} 

\subsection{Ablation Studies}
This section investigates the impact of different pretraining methods and the model's design.
%Specifically, we analyze the Transformer Encoder, which has self-trained on large-scale documents, and the CLIP text encoder used in our model; Our model's components: MLP layers and shallow, deep prompt method; Prompt template: different sentence template, single sentence template or independent attributes. The baseline model, illustrated in Figure \ref{framework}, utilizes the pre-trained transformer for the DiT single vision model, the CLIP text encoder, along with six different types of sentence templates, the deep prompt method, and the Cascade Mask-RCNN detection head. \\ \\

\begin{table}[]
	\caption{Ablation study for pretraining methods.}
	\centering
	\begin{tabular}{c|ccc}
		\hline
		Tag	& Method        & mAP    & $\Delta$               \\
		\hline
		%	Baselines w/o pretraining & 70.66	\\				 
		(a) & PromptDLA     & 78.69  &                       \\
		(b) & w/o pretraining & 70.66  & -8.03                \\
		(c) & \revise{VLP-pretraining(Layoutlmv3)} & \revise{76.38} & -2.31                \\
		(d) & w/o CLIP text encoder       & 76.33 & -2.36                \\
		\hline
	\end{tabular}
	\label{exp:model-init}
\end{table}

\noindent\textbf{Effect of Pretraining Methods}. We analyze the model's accuracy in three situations: without pretraining on document images,  with self-pretraining on document images using the single vision model, and with the DiT model being self-pretrained using a vision-language approach. Our baseline uses the pre-trained transformer encoder of DiT and the text encoder of CLIP. Compared to the baseline, the transformer encoder without pretraining reduces the mAP by 8.03, as illustrated in Table~\ref{exp:model-init}. Therefore, self-pretraining on large-scale document data can significantly enhance model accuracy by allowing the model to learn the generative document image representation via self-pretraining. Furthermore, the multi-modal pre-trained model from LayoutLMv3 is less effective than the single-model self-pre-trained transformer from DiT. Therefore, the single vision model's self-pre-training method is most suitable for dealing with layout analysis problems.
The CLIP text encoder is well-pretrained and can provide a prior textual representation. Table~\ref{exp:model-init} demonstrates that removing the CLIP text encoder and randomly initializing the prompt vector decreases 2.36 in mAP. The CLIP text encoder is already trained on large-scale image-text pairs. Thus, it can provide domain prior knowledge effectively.

\noindent\textbf{Effect of Prompt Location}. 
We explore the impact of prompt location in the transformer encoder, employing two distinct prompt methods illustrated in Fig.~\ref{deep-shallow}: Shallow Prompted Encoder (SPE) and Deep Prompted Encoder (DPE). SPE exclusively incorporates a prompt at the first layer of the transformer encoder, whereas DPE integrates a prompt at every layer. As shown in Table~\ref{exp-struc}, both SPE and DPE models exhibit improvements compared to the baseline model without a prompt. Notably, the DPE method surpasses the baseline by 2.26 mAP.
In the case of DPE, the CLIP text encoder generates 512-dimensional vectors, requiring MLP projection to the same 768-dimensional feature space as the image patch. We explore the design of the MLP for each transformer layer, specifically whether to use a shared MLP layer. Our observations indicate that the DPE method with a shared MLP layer results in an improvement of 2.01.

\begin{table}[h]
	\caption{Ablation study for prompt location.}
	\centering
	\begin{tabular}{c|ccc}
		\hline
		Tag&	Method     & mAP   & $\Delta$                \\
		\hline
		(a)&	Baselines   & 76.43 & \\
		(b)&	SPE         & 78.41 & +1.98\\
		(c)&	DPE 		& \textbf{78.69} & \textbf{+2.26}\\
		(d)&	DPE with share MLP Layer & 78.44 & +2.01\\
		%		PromptDLA  & 78.69 &                         \\
		%		Shallow Prompted Encoder  & 78.41 & -0.28    \\
		%		Share MLP Layer & 78.44 & -0.25             \\
		\hline
	\end{tabular}
	\label{exp-struc}
\end{table}

\noindent\textbf{Effect of Prompt Design}.
This study aims to investigate how to design prompts for document domain priors, exploring the correlation between prompts and DLA model performance on DocLayNet.
Table~\ref{exp-prompt} presents various prompt templates in the first column, the zero-shot document classification accuracy via CLIP in the second column, and the DLA results in the third column. The experiments demonstrate that prompts achieving higher accuracy in the CLIP classification task also lead to superior performance in our DLA model. This validates the rationale of our approach to creating prompt sets, focusing exclusively on the top-k accuracy rankings for the CLIP zero-shot classification task.
% Moreover, we observe that the more effective the prompt used, the better the zero-shot performance gets.

%This study draws inspiration from CLIP's prompt engineering and ensembling approach and aims to investigate the representation of textual descriptions for the document domain. The study first utilizes independent words to describe the domain of the document image. Second, the study employs the same sentence template for each domain. Finally, the study assigns different sentence templates to each domain as a baseline. The DocLayNet dataset comprises six domain document images, and therefore, the study's baseline employs six different sentence templates. As demonstrated in Table~\ref{exp-prompt}, using independent words to describe the domain of the document image results in a decrease in mAP from 78.69 to 78.51, and using the same template results in a reduction of mAP from 78.69 to 77.97 when compared to the baseline.

\begin{table}[h]
	\centering
 	\caption{Ablation study for different domain prompts.}
	\resizebox{\linewidth}{!}{
		\begin{tabular}{c|cc}
			\hline Prompt & Accuracy & mAP \\
			\hline 
			w/o prompt  &-   & 76.43 \\
			A page comes from \{Domain Class\} &  44.05& 77.49 \\
			A document page of \{Domain Class\} & 44.05 & 77.53\\
			A piece of paper concerning with \{Domain Class\} & 44.44 &77.55 \\
			A piece of paper comes from \{Domain Class\} & 46.99 &77.97 \\
			A document page comes from \{Domain Class\} & 48.45  &78.12 \\
			\hline
	\end{tabular}}
	\label{exp-prompt}
\end{table}

\begin{table*}[htbp]
	\centering
	\setlength{\tabcolsep}{1.5pt}
	\renewcommand{\arraystretch}{1}
	\caption{Experiments result on each document type in DocLayNet (“w/o” and “w” denote DiT baseline and PromptDLA, respectively).}
	\tiny  % ↓↓↓ Everything below will use a smaller font
	\resizebox{0.86\textwidth}{!}{
		\begin{tabular}{c|cccccccccccc|c}
			\hline  & & Caption & Footnote & Formula & List-item & Page-footer & Page-header & Picture & Section-header & Table & Text & Title & mAP \\
			\hline 
			\multirow{2}{*}{\begin{tabular}[c]{@{}c@{}}Finacial\\ reports\end{tabular}}	& w/o & \textbf{55.7} & 10.6 & - & 69.4 & 52.6 & 46.1 & 64.0 & 64.4 & 90.4 & 85.1 & 54.5 & 59.3 \\
			& w & 55.4 & \textbf{23.9} & - & \textbf{73.8} & \textbf{60.2} & \textbf{55.4} & \textbf{67.7} & \textbf{67.6} & \textbf{91.4} & \textbf{86.2} & \textbf{60.5} & \textbf{64.2} \\
			\hline
			\multirow{2}{*}{\begin{tabular}[c]{@{}c@{}}Government\\ tenders\end{tabular}} & w/o & 15.0 & 82.0 & 65.3 & 90.0 & \textbf{51.5} & 75.6 & 75.3 & 78.0 & \textbf{95.4} & 85.1 & 39.1 & 68.4 \\
			&w & \textbf{27.1} & \textbf{88.3} & \textbf{100.0} & \textbf{91.8} & \textbf{50.9} & \textbf{80.4} & \textbf{77.8} & \textbf{81.7} & 95.0 & \textbf{88.3} & \textbf{61.3} & \textbf{76.6} \\
			\hline
			\multirow{2}{*}{{\begin{tabular}[c]{@{}c@{}}Laws and\\ regulations\end{tabular}}} & w/o  & 28.0 & 91.6 & 16.7 & 81.6 & \textbf{42.9} & \textbf{71.8} & 49.5 & 68.0 & \textbf{70.4} & 84.1 & 84.4 & 62.6 \\
			& w & \textbf{28.5} & \textbf{96.1} & \textbf{18.5} & \textbf{82.0} & 42.4 & 64.4 & \textbf{50.4} & \textbf{69.5} & 67.7 & \textbf{84.7} & \textbf{84.6} & 62.6 \\
			\hline 
			\multirow{2}{*}{ Manuals } & w/o & 85.7 & 33.8 & - & 82.0 & 76.1 & 87.6 & 76.7 & 78.3 & 70.3 & 83.5 & 59.4 & 73.3 \\
			& w & \textbf{89.5} & \textbf{43.4} & - & \textbf{82.5} & \textbf{77.3} & \textbf{91.9} & \textbf{77.9} & \textbf{81.4} & \textbf{71.7} & \textbf{85.9} & \textbf{69.2} & \textbf{77.1} \\
			\hline 
			\multirow{2}{*}{ Patents } & w/o  & 78.0 & - & \textbf{61.6} & 90.4 & 85.8 & 91.1 & \textbf{89.8} & 91.8 & \textbf{92.9} & 93.1 & \textbf{88.5} & \textbf{86.3} \\
			&w & \textbf{81.1} & - & 51.8 & \textbf{91.7} & \textbf{87.3} & \textbf{92.0} & 87.7 & \textbf{91.9} & 92.8 & \textbf{93.6} & 86.6 & 85.7 \\
			\hline 
			\multirow{2}{*}{ \begin{tabular}[c]{@{}c@{}}Scientific\\ articles\end{tabular} } & w/o& 92.9 & 70.5 & 69.8 & 94.6 & 87.4 & 87.3 & 93.9 & 90.0 & \textbf{98.3} & 91.8 & \textbf{96.2} & 88.4 \\
			&w & \textbf{94.2} & \textbf{81.2} & \textbf{74.6} & \textbf{95.4} & \textbf{90.2} & \textbf{89.9} & 93.9 & \textbf{90.8} & 98.2 & \textbf{92.8} & 96.0 & \textbf{90.7} \\
			\hline
		\end{tabular}
	}
	\label{doclaynet-domain-result}
\end{table*}

\subsection{Discussions}

\noindent\textbf{Results on Each Document Type in DocLayNet.}
%As presented in Table~\ref{doclaynet-domain-result}, we trained PromptDLA and DiT using DocLayNet's entire training dataset and evaluated their performance separately on each document type in the test dataset. The results show that  PromptDLA outperformed DiT in terms of mAP, except for Laws and Regulations and Patents. However, categories such as "Page-header" and "Table" in Laws and Regulations and Patents, as well as Formula, Title, and Picture in Patents may not contain enough domain-specific information for accurate detection. Thus, the Prompt is ineffective in these categories.
As illustrated in Table~\ref{doclaynet-domain-result}, we trained PromptDLA and DiT using DocLayNet's entire training dataset and evaluated their performance separately on each document type in the test dataset. Compared to the baseline DiT model without any domain-aware prompt, the model with Prompt outperforms it in terms of mAP, except for "Laws and Regulations" and "Patents". Despite a decrease in our model's mAP in the "Patents" domain, it still surpassed the baseline model in most categories, such as "Caption" and "Page-header." We attribute the decline in the Formula and Picture categories to the limited correlation between Formula and Picture in the field of patents.

% \noindent\textbf{Effectiveness of CLIP on Document}.
% The motivation behind using CLIP as a prompter is its impressive ability to learn visual and textual representations. In this section, we validate CLIP's ability to provide prior knowledge for documents through zero-shot document classification.  Excitingly, as shown in Table~\ref{zero-shot classify}, CLIP achieves 48.45\% accuracy across $6$ different domains in DocLayNet and 54.55\% accuracy across $7$ different domains in M6Doc. This indicates that CLIP possesses prior information about th andayout image.

\noindent\textbf{Validating CLIP's Prior Knowledge of Documents.}
The motivation behind using CLIP as a prompter is its impressive ability to learn visual and textual representations. We validate our choice of CLIP as a prompter by evaluating its zero-shot document classification performance. As shown in Table~\ref{zero-shot classify}, CLIP achieves 48.45\% accuracy on DocLayNet and 54.55\% on M6Doc without any fine-tuning. This indicates that CLIP possesses prior information about the document layout image.

%\noindent\revise{\textbf{Computational Overhead.} To evaluate the tradeoffs between the speed and accuracy of our PromptDLA,  we test the PromptDLA with different pre-trained DLA models on various datasets. As shown in Table~\ref{time-cost}, the average inference time on DiT is 6.75 FPS, while with the PromptDLA, it is 6.62 FPS with only 0.13 FPS decrease.} % Thus, we can conclude that the PromptDLA can perform better with a slight time cost. 
\noindent\textbf{Computational Overhead.} PromptDLA is computationally efficient, adding negligible overhead. As shown in Table~\ref{time-cost}, applying our method to a DiT backbone reduces inference speed by a mere 0.13 FPS (from 6.75 to 6.62 FPS) on an RTX 3090 GPU. This trend holds across various models.

\begin{table}[h]
\centering
\caption{Zero-shot document classification performance of CLIP.}
\begin{tabular}{cccc}
\hline 
      &DocLayNet & M6Doc\\
\hline 
\#Document Types & 6 & 7 \\
\hline Accuracy & 48.45\% & 54.55\%  \\
\hline
\end{tabular}
\label{zero-shot classify}
\end{table}

\begin{table}[h]
	\setlength{\tabcolsep}{1.5pt}

	\caption{Computational overhead of PromptDLA.  }
    \centering
	\resizebox{0.7\linewidth}{!}
	{
		\begin{tabular}{cccc}
			\hline Model & mAP@IOU[0.50:0.95]  & FPS \\
			\hline
			DiT-Base & 76.4  & 6.75 \\
			\quad \quad +PromptDLA & 78.7(+2.3) & 6.62(-0.13) \\
			\hline 
			LayoutLMv3 & 75.7  & 4.44 \\
			\quad \quad +PromptDLA & 76.4(+0.7) & 4.41(-0.03) \\
			\hline
	\end{tabular}}
	\label{time-cost}
\end{table}

\noindent\textbf{Visualization}.
Fig.~\ref{vi DocLayNet result} presents the visualized results on the Financial Reports and Laws domain from DocLayNeta comparison between ground truth, DiT, and PromptDLA is presented. For the sample of Financial in the 1st row, DiT misidentifies the background as "Figure" and recognizes the whole "Table" as two separates. At the same time, the PromptDLA removes the misclassification of "Figure" and produces a precise box of "Table." Moreover, the sample of Laws in 2nd row shows that our method precisely excludes the text box and identifies only the text inside it when processing manuals with text boxes. In contrast, DiT incorrectly identifies the text box as a complete figure. These qualitative results demonstrate the ability of the PromptDLA to recognize ambiguous objects by domain prior. Besides,  Fig.~\ref{more vi DocLayNet results} in Appendix shows more visualization examples. %\revise{ (We provide additional visualization examples at the end of the paper, shown in Figure~\ref{more vi DocLayNet results}}. 

\begin{figure*}[h]
	\centering
	\includegraphics[width=0.88\linewidth]{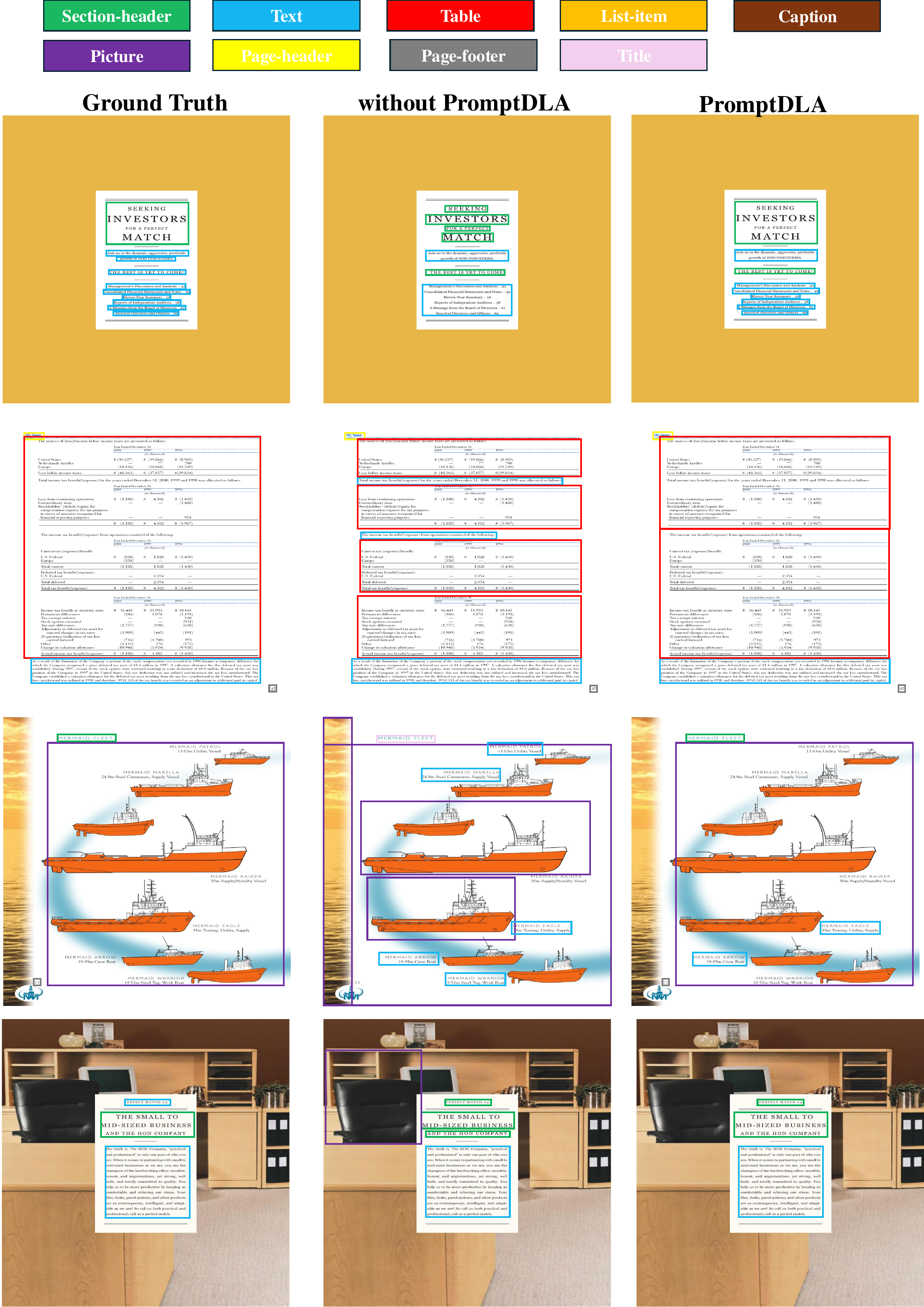}
	\caption{Appendix: More visualization examples on the DocLayNet dataset. From left to right, columns show Ground Truth annotations, predictions without PromptDLA, and predictions with PromptDLA, highlighting document layout segmentation accuracy improvements.}
	\label{more vi DocLayNet results}
\end{figure*} 

\section{Conclusions}
We propose a novel PromptDLA framework, which can explicitly introduce domain prior into the DLA frameworks and steer DLA automatically, distinguishing the variability of different domains. The PromptDLA features a unique domain-aware prompter that could customize prompts according to the specific attributes of the data domain. We underscore the significance of utilizing domain priors in DLA through extensive experiments. The results show a new state-of-the-art performance across multiple datasets, including DocLayNet (78.7),  M6Doc (69.2), and D$^4$LA (69.1). It's worth mentioning that the proposed domain-aware prompter is easily plugged into enhance different DLA frameworks.{ While PromptDLA demonstrates strong performance and adaptability, several avenues for future work remain. A key direction is efficiency optimization. Integrating large language or vision-language models, particularly in the prompter component, introduces computational overhead compared with baseline DLA models.}

\bibliographystyle{IEEEtran}
\bibliography{main}
\end{document}